\documentclass{article} 

\usepackage{amsmath,amsfonts}
\usepackage{algorithm}
\usepackage{algorithmic}
\usepackage{array}
\usepackage[caption=false,font=normalsize,labelfont=sf,textfont=sf]{subfig}
\usepackage{textcomp}
\usepackage{stfloats}
\usepackage{url}
\usepackage{verbatim}
\usepackage{graphicx}
\usepackage{cite}
\usepackage{mathtools} 
\usepackage{siunitx} 
\usepackage{booktabs} 
\usepackage{tikz} 
\usepackage{amsthm}
\usepackage{booktabs}
\usepackage{xcolor}
\usepackage{colortbl}
\usepackage{bbold}
\usepackage{listings}
\usepackage{bm}
\usepackage{fancyhdr}
\usepackage{hyperref}
\usepackage{wrapfig}

\newcommand{\walon}[1]{{\color{black}#1}\normalfont}
\newcommand{\iclr}[1]{{\color{black}#1}\normalfont}
\newcommand{\iclrA}[1]{{\color{black}#1}\normalfont}

\title{Mitigating Forgetting in Online Continual Learning via Contrasting Semantically Distinct Augmentations}

\author{Sheng-Feng Yu$^{\dag\ddag}$ \hfill Wei-Chen Chiu$^{\dag}$  \\
$^{\dag}$National Yang Ming Chiao Tung University, Taiwan \\
$^{\ddag}$Macronix International Co., Ltd.\\
\texttt{robertyu1@mxic.com.tw \hfill walon@cs.nctu.edu.tw} \\
}

\begin{document}

\maketitle

\begin{abstract}
Online continual learning (OCL) aims to enable model learning from a non-stationary data stream to continuously acquire new knowledge as well as retain the learnt one, under the constraints of having limited system size and computational cost, in which the main challenge comes from the ``catastrophic forgetting'' issue -- the inability to well remember the learnt knowledge while learning the new ones. 
With the specific focus on the class-incremental OCL scenario, i.e. OCL for classification, the recent advance incorporates the contrastive learning technique for learning more generalised feature representation to achieve the state-of-the-art performance but is still unable to fully resolve the catastrophic forgetting.
In this paper, we follow the strategy of adopting contrastive learning but further introduce the \textit{semantically distinct augmentation} technique, in which it leverages strong augmentation to generate more data samples, and we show that considering these samples semantically different from their original classes (thus being related to the out-of-distribution samples) 
in the contrastive learning mechanism 
contributes to alleviate forgetting and facilitate model stability. 
Moreover, in addition to contrastive learning, the typical classification mechanism and objective (i.e. softmax classifier and cross-entropy loss) are included in our model design for \iclr{
utilising the label information}, but particularly equipped with a sampling strategy to tackle the tendency of favouring the new classes (i.e. model bias towards the recently learnt classes).
Upon conducting extensive experiments on CIFAR-10, CIFAR-100, and Mini-Imagenet datasets, our proposed method is shown to achieve superior performance against various baselines.
\end{abstract}

\section{Introduction}
The ability to continually learning new knowledge is getting more and more important for machine learning models nowadays as the increasing demands of automation and the dynamic nature of our environment, e.g. the visual recognition system of the goods in the intelligent self-checkout system for smart retail should be able to classify the newly-added items or the existing items with new packing. In particular, the model cannot be trained from scratch whenever the classes or recognition targets increase. Instead, it needs to keep adapting
itself to learn new knowledge on the fly over time. \textit{Online continual learning} (OCL)~\cite{delange2021survey} is one of the 
topics getting popular these years to serve such purpose, where the machine learning agent continually learns a few new concepts every once in awhile without forgetting the others (i.e. what the agent has learnt previously). 

If the agent continually learns to classify a new set of unseen classes, this particular problem is named \textit{class-incremental OCL}. It is one of the most prevalent settings in the community of OCL. However, learning on unseen classes would change the model parameters (and the feature representation space) optimised for the old classes. Hence, the model classification accuracy on the old classes inevitably deteriorates. This phenomenon is well-known and called \textit{Catastrophic Forgetting}~\cite{mccloskey1989catastrophic}.

\begin{wrapfigure}{R}{0.45\textwidth}
\vspace{-2\baselineskip}
\centering
\subfloat[Common]{\includegraphics[width=1.2in]{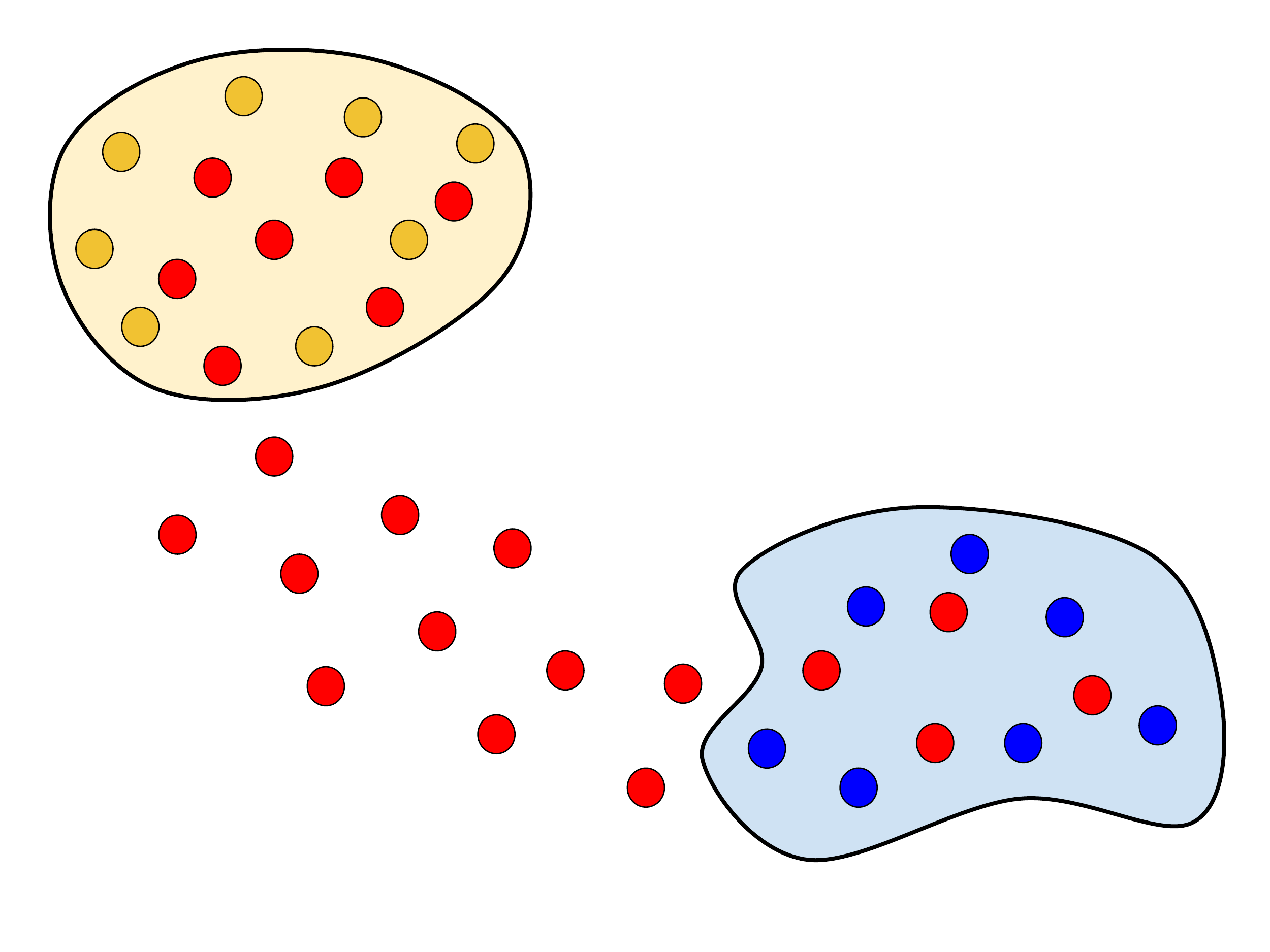}\label{fig:bad}}
\hfil
\subfloat[Generalisable]{\includegraphics[width=1.2in]{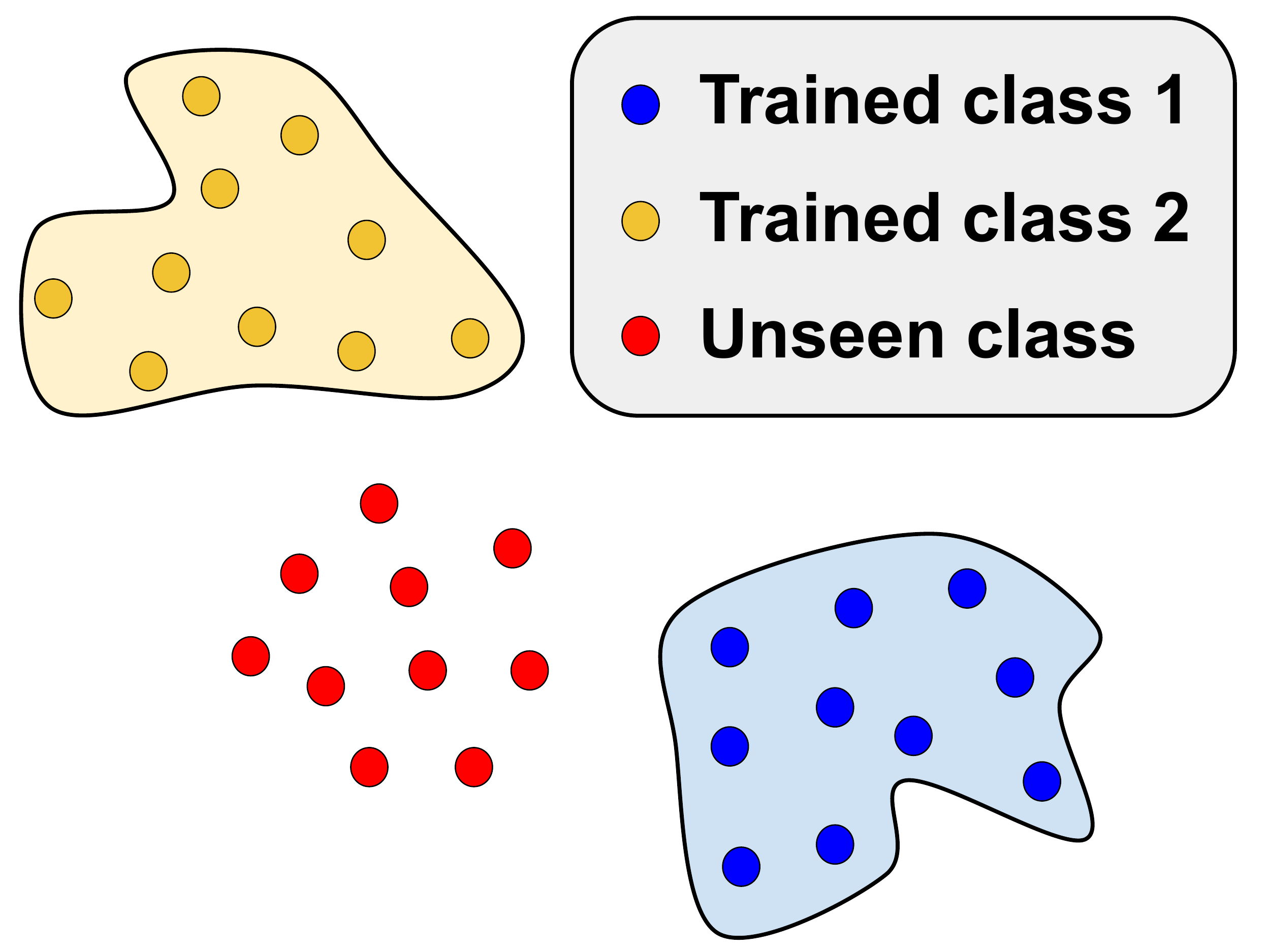}\label{fig:good}}
\caption{The main motivation of this work comes from the idea of learning a generalisable representation. The generalisable feature representation would ideally make the unseen classes more separable from the seen classes. The representation thus requires less adjustment while learning the new classes, and such a property naturally reduces the forgetting phenomenon as the main portion (responsible for feature extraction) of the classification model is minorly updated during the incremental learning procedure.}
\label{fig:representation}
\vspace{-\baselineskip}
\end{wrapfigure}


To address this issue,~\cite{chaudhry2019er} propose using a small memory to store the learnt examples. When learning new classes, the model is retrained/updated by using not only the recently received samples that belong to new classes but the stored examples from old classes, in which such a strategy attempts to maintain the accuracy for both the old and new classes. 
%
Another widely-adopted idea comes from~\cite{kirkpatrick2017ewc}, where they propose to regularise the model learning by constraining the update of important parameters in order to alleviate the catastrophic forgetting. 
However, limiting the model update space deteriorates the learning ability of the model. 
%
Recently, the introduction of learning a generalisable representation~\cite{mai2021scr,cha2021co2l,gallardo2021self,pham2021dualnet} brings another break.
Basically, learning the generalisable representation aims to not only distinguish between the learnt/seen classes but also have higher intrinsic dimension such that the features from unseen classes are more likely to be distributed away from the seen ones (cf. Figure \ref{fig:representation}), i.e. the feature representations are richer and more reusable/transferable for the unseen classes in which the model requires relatively minor adjustments to achieve high accuracy for the new classes.

Among the three ideas described above, learning a generalisable representation has benefited from the recent advance of self-supervised representation learning~\cite{noroozi2016jigsaw,gidaris2018rot}. 
In particular, contrastive learning, \iclr{e.g. SimCLR~\cite{chen2020simclr} and SimSiam~\cite{simsiam}}, has shown its effectiveness in learning generalisable image representation.
For instance, SimCLR leverages the composition of multiple carefully chosen data augmentations, such as cropping and colour distortion, to generate random views, and the model learns to align these random views in the representation space with the ones of the same label\footnote{In the unsupervised setting for the contrastive learning, the image index is commonly treated as the label.}.

However, \cite{chen2020simclr} observed that some augmentations (e.g. rotation, noise) are too strong to deteriorate the representation quality if they involve in the view generation, and \cite{wang2021prelax} discovered that the semantic shift caused by these augmentations is too large to align the corresponding random views well.
Despite this, it does not imply that strong augmentations~\cite{noroozi2016jigsaw,gidaris2018rot} cannot provide meaningful semantics for representation learning. For example, \cite{gidaris2018rot} propose learning image representation by predicting rotation.
Hence, those augmentations can still be utilised for contrastive representation learning, but we should be aware of not encouraging the model to align the views with those strong augmentations which could cause the large semantic shift.
In turn, if the diverse views produced by the strong augmentation are treated as belonging to the other classes which are distinct from their original samples, additionally considering them in contrastive learning could help the model learn to extract rich features and represent the unseen classes better (according to the observation made by~\cite{hendrycks2019oe} where the diverse views together are similar to the auxiliary dataset and the unseen classes are analogous to the out-of-distribution samples).
In this work, we base on such a line of thought and propose a technique named \textit{semantically distinct augmentation (SDA)}: given a mini-batch composed of the training samples, the strong transformations are first applied to these training samples to produce diverse views which are treated as from different/novel classes (cf. Figure~\ref{fig:aug}) and are added back to the mini-batch, then both \iclr{contrastive learning and softmax classifier are} applied to the extended mini-batch (with having both the original samples and their corresponding diverse views). 
%
The \iclr{two-pronged benefits are introduced by such SDA technique: First, learning on a diverse dataset allows the model to get a more generalisable representation and mitigates catastrophic forgetting; Second, as data comes in a stream for the OCL setting such that each data sample ideally can only be adopted once for training, our SDA leverages strong augmentations for the attempt on making best use of every sample.
}

%
Furthermore, as online continual learning has a non-stationary data stream, the model is likely to face the imbalanced training set (where the training samples are mostly from the newly added classes), thus the softmax classifier would suffer from the class-imbalanced problem~\cite{kang2019decoupling}. We hence adopt a specifically-designed sampling strategy to balance the learning between old and new classes. With conducting experiments on several datasets and different settings of online continual learning, our full model equipped with all the aforementioned designs (named as \textbf{SDAF}) is demonstrated to provide the state-of-the-art performance in comparison to various baselines.

\section{Related Work}
\subsection{Online Continual Learning}
The goal of class-incremental online continual learning~\cite{chen2018lifelong} focuses on how an artificially intelligent agent learns to classify new classes without forgetting its knowledge on the classes previously learnt (where such an issue is the so-called catastrophic forgetting). To tackle against the catastrophic forgetting, one should balance the model learning between the old classes and new classes, where the literature roughly contains three branches: 

(1) \textbf{Experience Replay}. 
\cite{chaudhry2019er} suggests that the online learning agent equips a fixed-sized memory to store the learnt examples, then the model repeatedly replays the samples from the memory to alleviate catastrophic forgetting. In particular, they adopt the reservoir sampling~\cite{10.1145/3147.3165reservoir} strategy to draw the samples from the memory for model training, such strategy ensures the sampling result being equivalent to having uniform sampling from the stream data without knowing the sequence length.
Follow-up works~\cite{NIPS2019_mir,NEURIPS2019gss,shim2021aser} assume that every training data has different importance, and remembering a few critical samples is enough for keeping the data distribution. 
Nevertheless, \cite{prabhu2020gdumb} show that the simple greedy selection (i.e. storing the latest examples) is actually able to achieve comparable performance without any cost to evaluate the importance of samples.

(2) \textbf{Regularisation and Constraint Optimisation}. \cite{kirkpatrick2017ewc,li2017lwf,chaudhry2018ewcpp} regularise the network update to alleviate catastrophic forgetting during learning new classes, and these approaches are often efficient and usually have little extra cost. For instance, \cite{lopez2017gem,AGEM} constrain the model optimisation such that the loss on past classes never increases. However, both works limit the space for the model optimisation, and hence they suffer from the inability of learning new classes. 

(3) \textbf{Improving Representation Learning}.
Online aware meta-learning (OML)~\cite{NEURIPS2019oml} uses a meta-learning objective to learn the sparse representation which easily adapts to new classes to mitigate catastrophic forgetting.
\iclr{And other approaches instead aim to learn the generalisable representations~\cite{mai2021scr,pham2021dualnet,gu2022dvc}.} As the generalisable representation ideally should provide better support for not only the seen classes but also the unseen ones, hence it requires fewer tuning to optimise for the new classes and suffers less from forgetting.

There are other approaches that are unable to categorise into the above three branches. 
First, knowledge distillation~\cite{hinton2015distilling} keeps an old model as a teacher to preserve the learnt knowledge~\cite{li2017lwf,buzzega2020dark,cha2021co2l}.
Second, the expansion-based online continual algorithm dynamically expands the network capacity upon the arrival of new classes~\cite{yoon2018lifelong,Lee2020cndpm}.
We do not take these approaches in account here, as they need additional resources for computation (e.g. much more additional memory for storing the old model to perform knowledge distillation; and continuously growing model size for the expansion-based methods in which it means that the requirement of memory space also keeps increasing), and we only consider the methods with similar computational cost as our baselines to make comparison.

In this work, we focus on mitigating the forgetting issue by continually learning a generalisable representation. The learning system additionally equips a small memory for replay, and every sample only appears once in the entire training trajectory except that it is stored in the memory. 

\subsection{Image Representation Learning}
Image representation learning is an essential foundation for various computer vision tasks. Especially, self-supervised learning is one of the most thrilling branches in this field. Self-supervised learning encourages the machine to learn image representation from a pretext task, which is able to automatically generate a supervision signal via a predefined transformation without any human labelling. For example, image permutation and rotation prediction~\cite{noroozi2016jigsaw,gidaris2018rot} help the model learn the image feature representation.

In particular, recent works on self-supervised learning advance to contrast between images to perform representation learning by leveraging the combination of various transformations, called \iclr{contrastive learning~\cite{chen2020simclr, byol, simsiam, zbontar2021barlow, bardes2021vicreg}}. For instance, SimCLR~\cite{chen2020simclr} as a representative work first generates a pair of positive views by applying a sequence of transformations to an image twice,  
then it learns image representation by attracting the positive pairs and pushing negative views from other images away.
%
\iclr{
SimSiam~\cite{simsiam} can be thought of as ``SimCLR without negatives'', it introduces a predictor network in its forward process on one view and applies a stop-gradient operation in its backward process on the other view. Moreover, as it uses neither the negative sample pairs nor the momentum encoders (what other self-supervised methods, e.g. MoCO~\cite{he2019moco} and BYOL~\cite{byol}, would need),  
it has smaller model size during training (compared to MoCO and BYOL) as well as better support for the small training batch. SimSiam thus becomes suitable for the computational-cost-sensitive problems such as OCL, the main topic of this paper.
}


We argue that representation learning has a high potential to mitigate catastrophic forgetting by increasing the feature generalisability. And our method proposes to utilise a strong data augmentation to boost the feature generalisability learnt by the contrastive learning, which will be detailed later in the following sections. 


\section{Methodology}

\begin{figure*}[!t]
    \centering

    \setcounter{subfigure}{0}
    \subfloat[SCR]{\includegraphics[width=1.225in]{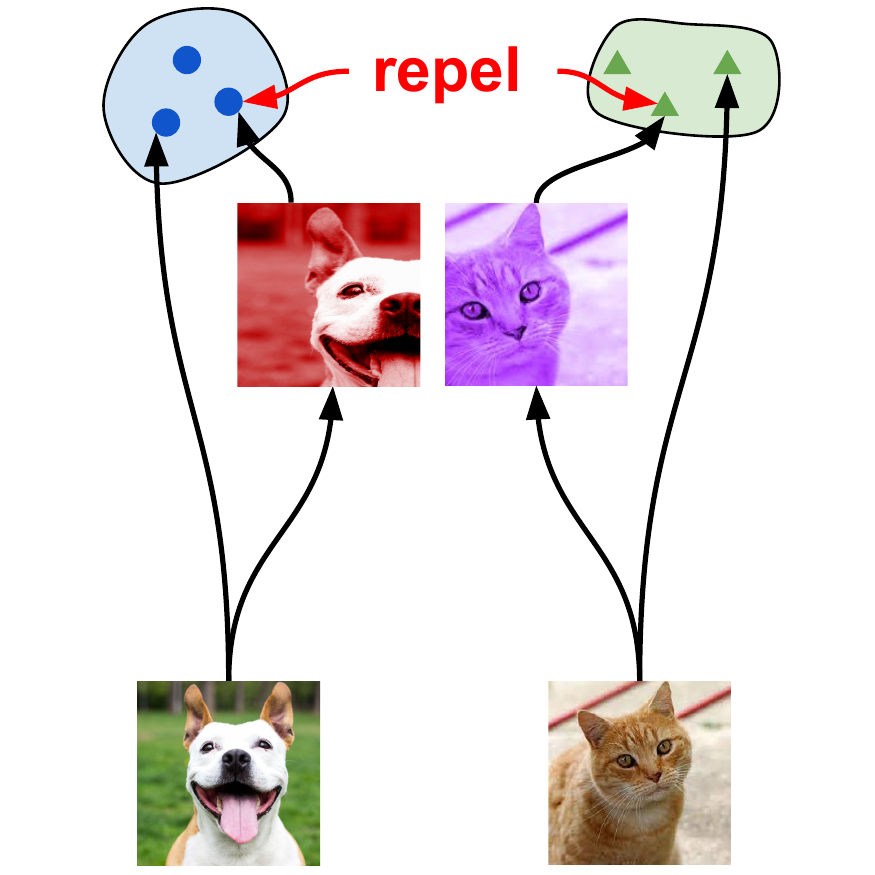}\label{fig:scr}}
    \hfil
    \subfloat[SCL]{\includegraphics[width=1.225in]{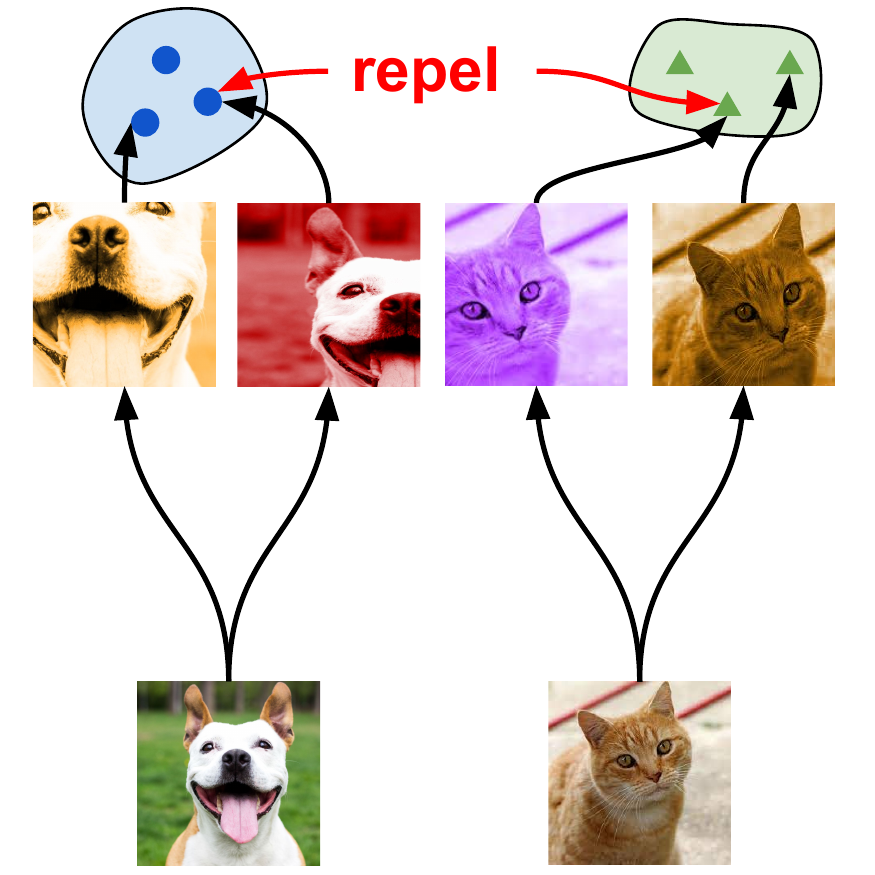}\label{fig:clr}}
    \hfil
    \subfloat[SDA]{\includegraphics[width=1.225in]{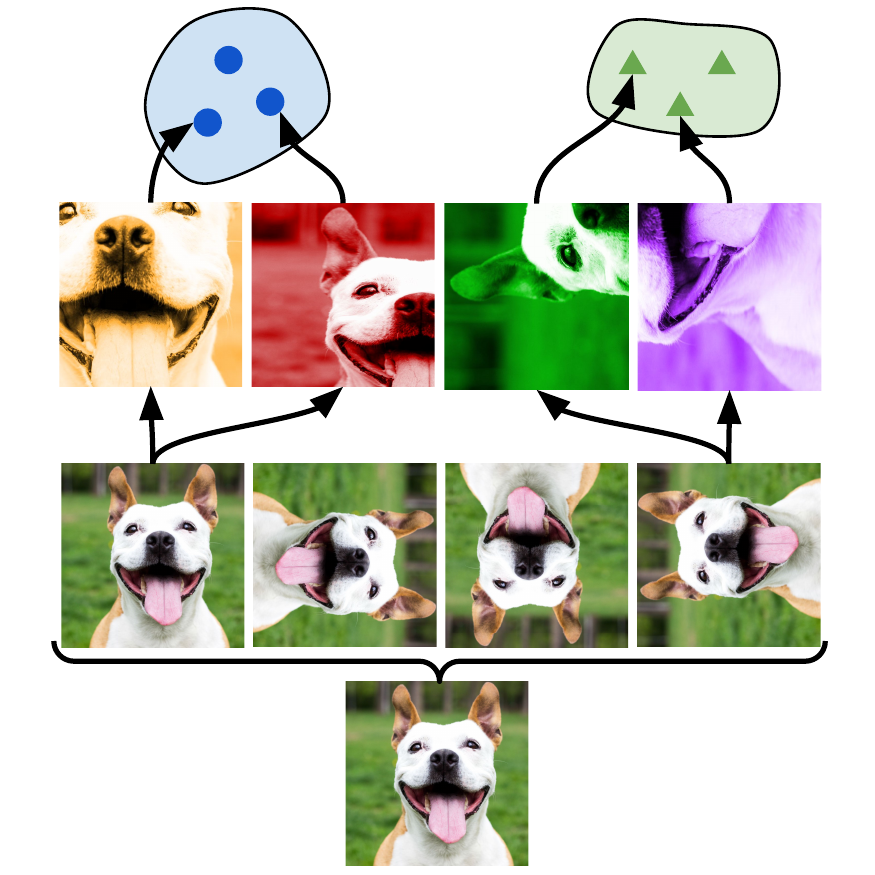}\label{fig:sda}}
    \hfil
    \subfloat{\includegraphics[width=1.05in]{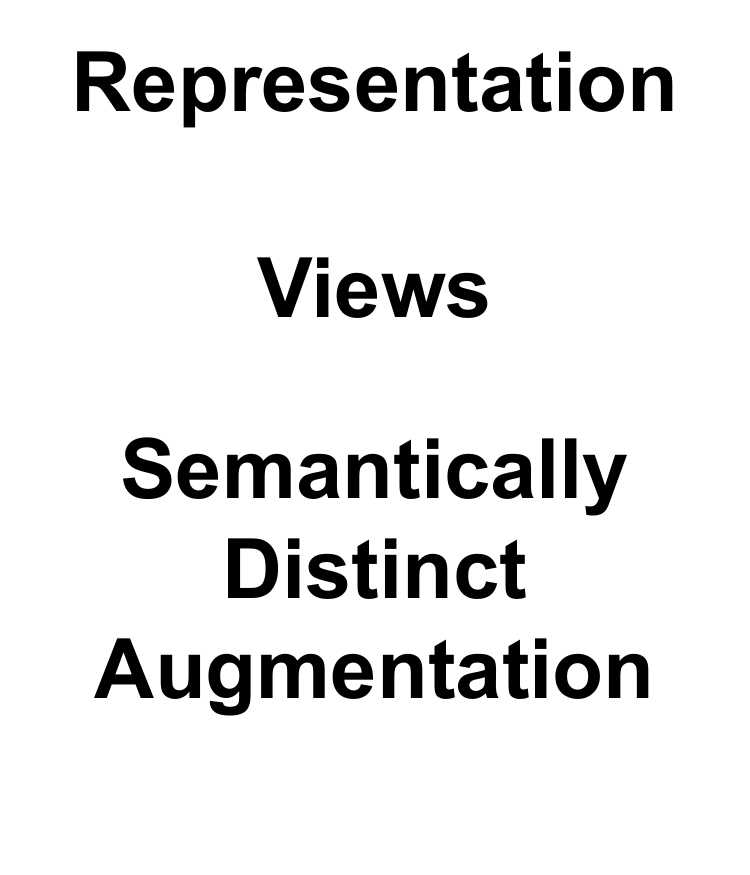}\label{fig:name}}
        \caption{\iclr{The comparison among \textbf{(a)} supervised contrastive replay (SCR), \textbf{(b)} supervised contrastive learning (SCL), and \textbf{(c)} our proposed SDA model. SCR contrasts the original image to a random view, and SCL contrasts between two random views. Our proposed SDA first augments an image into $K$ instances in which each instance is treated as belonging to different classes, then every instance generates two views. In results, there are $2K$ views in total. Specifically, since the $K$ instances augmented via SDA are now treated as $K$ novel classes, the classification scenario here is analogous to a $\{K(C_{old}+C_t)\}$-ways classification problem. Noting that, for every representation on the top, the points with the same colour are encouraged to form a compact group during the learning. Moreover, the repulsion between different classes is only performed in SCR/SCL but not used in our SDA (as SDA adopts the contrastive learning mechanism from SimSiam, cf. Section~\ref{sda}). }
        }
        \label{fig:aug}
\end{figure*}

In this work, given a machine learning agent to execute the class-incremental online continual learning, we assume that it has a fixed size memory $\mathcal{M}$ to store the exemplars of the old/learnt classes (for the purpose of experience replay) and there are $T$ training stages for the entire training process, where for each training stage $t \in \{1\cdots T\}$ the agent will learn from the training examples that arrive as a data stream \iclr{(in which it means that each training sample only appears once unless it is stored in the memory)} for recognising a set of new classes $\mathcal{C}_t$. Please note that in the following paragraphs we would misuse $\mathcal{M}$ to represent the experience replay memory or its size for simplicity. Following a similar setting as in previous works, the classes learnt at each training stage are assumed to be disjoint for simplicity, i.e. $\mathcal{C}_i \cap \mathcal{C}_j = \phi$ for any $i \ne j$, and we denote the training samples received during the training stage $t$ as $\mathcal{D}_t$ where they belong to the classes $\mathcal{C}_t$. As the target of class-incremental online continual learning at the training stage $t$ is to let the machine learning agent not only learn the new classes $\mathcal{C}_t$ but also maintain its recognition ability for the old classes learnt during previous stages, the learning scenario at the training stage $t$ is a $\{C_{\text{old}} + C_t\}$-ways classification problem where $C_{\text{old}}=\sum_{i=1}^{t-1} C_i$ and $C_i$ denotes the cardinality of $\mathcal{C}_i$. Without loss of generality, we index the old classes $\mathcal{C}_{\text{old}}$ by $\{ 1, 2, \dots, C_{\text{old}}\}$ and the new classes $\mathcal{C}_t$ by $\{ C_{\text{old}} + 1, C_{\text{old}} + 2, \dots, C_{\text{old}} + C_t\}$. In detail, the data stream at the training stage $t$ is composed of $U$ data batches
$\mathcal{B}_{t}^u$ where $u = 1\cdots U$, and each batch $\mathcal{B}_{t}^u$ contains a group of training samples $x_i$ and the corresponding class labels $y_i$ where $y_i\in\{C_{\text{old}}+1,C_{\text{old}}+2,\dots,C_{\text{old}}+C_t\}$.
These batches are disjoint, i.e. every training sample only appears among batches once during the data stream, \iclr{this setting is also called \textit{one epoch setting}.}

Without loss of generality, here we summarise a generic algorithmic procedure for the class-incremental online continual learning methods with the experience replay memory in Algorithm~\ref{alg:algorithm}. Basically, when the machine learning agent receives a data batch $B_{t}^u$ during the training stage $t$, it will use the data from $B_{t}^u$ (belonging to the new classes) as well as the samples from the experience replay memory (mostly belonging to the old classes) to train itself for $I$ iterations (for simplicity, we assume that $I$ is a constant). \iclr{Please note that, we just train the same sample for $I$ times which does not violate the one epoch setting}. In particular, at each iteration, the agent utilises the data $B_{t}^u \cup B_M$ to perform the training where $B_M$ denotes the $m$ samples retrieved from the experience replay memory. After $I$ iterations of training based on the batch $B_{t}^u$ and the memory, the MemoryUpdate operation is performed to replace some exemplars in the memory $\mathcal{M}$ with the ones sampled from $B_{t}^u$, 
where the reservoir sampling algorithm~\cite{10.1145/3147.3165reservoir} is adopted for such a MemoryUpdate operation in our work following the practise in \cite{mai2021scr}.

\begin{wrapfigure}{R}{0.5\linewidth}
\vspace{-2\baselineskip}
\begin{minipage}{\linewidth}
\begin{algorithm}[H]
\caption{Generic class-incremental online continual learning algorithm with the experience replay memory}
\label{alg:algorithm}
\textbf{Input}: 
Learning rate $\alpha$;
        The number of iterations for SGD update $I$; Training objective $\mathcal{L}$\\
\textbf{Parameter}: $\theta$

\begin{algorithmic}[1] 
\STATE Memory $\mathcal{M} \leftarrow \{\}$
\FOR{$t=1 ~  \text{to} ~ T$}
    \WHILE{$B_{t}^u \sim D_t$}
        \FOR{$i=1 ~ \text{to} ~ I$} 
                \STATE $B_M \leftarrow \text{MemoryRetrieval}(\mathcal{M}, m)$
            \STATE $\theta \leftarrow$ SGD$(B_{t}^u \cup B_M, \mathcal{L}, \theta, \alpha)$
        \ENDFOR
        \STATE $\mathcal{M} \leftarrow \text{MemoryUpdate}(B_{t}^u, \mathcal{M})$
    \ENDWHILE
\ENDFOR
\RETURN $\theta$
\end{algorithmic}

\end{algorithm}
\end{minipage}
\end{wrapfigure}

\subsection{Preliminary}
As motivated previously that our proposed method stems from the idea of learning a generalisable representation via contrastive learning, where a recent work~\cite{mai2021scr} adopts such idea to achieve the state-of-the-art in class-incremental OCL,
here we hence review several key references (e.g. SimCLR~\cite{chen2020simclr} and SCL~\cite{khosla2020supervised}) for traversing the main ideas behind~\cite{mai2021scr} in order to build the preliminary of our method.

First, SimCLR~\cite{chen2020simclr}, as a representative approach of contrastive learning, consists of three components: a random transformation module $\mathcal{H}$, an encoder network $\mathcal{F}$, and a projection network $\mathcal{G}$. Basically, the transformation module $\mathcal{H}$ adopts a sequence of random transformations (e.g. sequentially applying random crop, \iclr{random horizontal flip}, random colour distortion, and random gray-scaling) to generate different views $\tilde{x}_i^{(j)}$ for every image $x_i$ in a training batch $\{x_i\}_{i=1 \dots \mathbb{B}}$, where $\mathbb{B}$ is the number of samples in a batch, $\tilde{x}_i^{(j)}=H^{(j)}(x_i)$, and the transformation operation $H^{(j)}$ is re-sampled from $\mathcal{H}$ for each $x_i$. Based on such random transformations, we construct a set of views $\tilde{x}_{i}^{(j)}$ together with their corresponding image index $i$:
\begin{equation}
\mathbf{V}=\bigcup\limits_{i=1}^\mathbb{B}\bigcup\limits_{j=1}^{2}\{(\tilde{x}_{i}^{(j)},i)\mid \tilde{x}_{i}^{(j)}=H^{(j)}(x_i)\}
\end{equation}
With denoting $z_{i}^{(j)}=\mathcal{G}(\mathcal{F}(\tilde{x}_{i}^{(j)}))$, the goal of SimCLR training is to learn the feature encoder $\mathcal{F}$ via the objective of encouraging two vectors $z_{i}^{(j)}$ and $z_{i}^{({j}')}$ obtained from the same $x_i$ but under different transformations (i.e. positive pairs) to attract each other while enforcing the $z$ vectors to repel once they originate from different images (i.e. negative pairs). 

In comparison to SimCLR which is self-supervised as the positive and negative pairs are simply determined by their corresponding image indexes, Supervised Contrastive Learning (SCL)~\cite{khosla2020supervised} takes the class labels $y$ into consideration thus being supervised, in which the training views are constructed by:
\begin{align}
\mathbf{V}_{\text{SCL}} = \bigcup\limits_{i=1}^\mathbb{B}\bigcup\limits_{j=1}^2\{(\tilde{x}_{i}^{(j)}, y_{i})\mid \tilde{x}_{i}^{(j)}=H^{(j)}(x_i)\}
\end{align}
where the transformation module $\mathcal{H}$ here is the same as the one used in SimCLR. Similarly, given the training views $\mathbf{V}_{\text{SCL}}$, the objective of SCL is to encourage the attraction between $z$ vectors from the same class and enforce the repulsion between the ones from different classes, for learning the feature extractor $\mathcal{F}$.

Supervised contrastive replay (SCR)~\cite{mai2021scr} adapts SCL for the problem of class-incremental online continual learning and provides the state-of-the-art performance, where the main difference between SCR and SCL comes from the transformation module, as visualised in Figure~\ref{fig:scr} and Figure~\ref{fig:clr} respectively: For SCL, its transformation module $\mathcal{H}$ applies two distinct transformations on the input image to construct the positive pair, while a positive pair in SCR is built upon an input image $x_i$ and its random view $\tilde{x}_i$, thus SCR in general has lower randomness than SCL. With denoting the original images as $\mathbf{V}_{\text{ori}}=\{(\tilde{x}_{i}, y_{i}) \mid \tilde{x}_{i}=x_i\}$, the training views for SCR are then formed by:
\begin{equation}
\mathbf{V}_{\text{SCR}} = \bigcup\limits_{i=1}^\mathbb{B} ( \{(\tilde{x}_{i},y_{i}) \mid \tilde{x}_{i}=H(x_i)\} 
\cup \mathbf{V}_{\text{ori}})
\end{equation}
In particular, when SCR and SCL are both applied in the class-incremental OCL scenario, as SCR has lower randomness than SCL during the constructing training views, it is more likely to provide higher classification accuracy than SCL in the first few training stages; however, such lower randomness of SCR in turn sacrifices the potential for learning more diverse (thus more generalised) representations hence leading to lower classification accuracy of SCR with respect to SCL in the later training stages. \iclrA{Please refer to Appendix~\ref{stagewise_acc} for more details.}

\subsection{Semantically Distinct Augmentation} \label{sda}
The performance difference versus training stages caused by the aforementioned randomness between SCR and SCL motivates us to conduct further research on the impact of the random transformations upon the representation learning. As found by the work of SimCLR~\cite{chen2020simclr}, adding some particular transformations (e.g. rotation or blur) into the transformation module would instead hurt the quality of learnt representations as these transformations cause more significant distortion to the input image (thus having the semantic shift). Following such an empirical observation, we propose the mechanism named \textit{Semantically Distinct Augmentation (SDA)} which is applied on the input image $x$ before the transformation module $\mathcal{H}$.
The SDA consists of multiple deterministic augmentations, and every augmentation would cause a distinct semantics change of the input images. In results, if SDA are adopted during the representation learning, the feature space tends to have higher intrinsic dimension (which leads to more generalised features) for handling the diverse semantics produced by SDA.
Later in experiments, we demonstrate that using such an SDA mechanism benefits the online continual learning to learn more generalised representations, thus leading to superior performance.

In detail, we assume that there are $K$ strong deterministic increases $\mathcal{S}=\{S_1, S_2, \dots, S_K \}$ in the SDA mechanism, and every augmentation in $\mathcal{S}$ applies to each sample $x_i$, that is $S_k (x_i)$. Then, similarly to SCL, $H^{(j)}$ is used to generate random views, $\tilde{x}_{ik}^{(j)}=H^{(j)}(S_k (x_i))$, as visualised in Figure \ref{fig:sda}, and the extended label space is defined by:
\begin{equation}
\tilde{y}_{ik}=K(y_i-1)+k
\end{equation}
where the original label $y_i$ extends to $K$ different classes.
Based on such an extended label space, the batch of views for training is defined as:
\begin{equation}
\mathbf{V}_{\text{SDA}} = \bigcup\limits_{i=1}^\mathbb{B}\bigcup\limits_{j=1}^2\bigcup\limits_{k=1}^K\{(\tilde{x}_{ik}^{(j)}, \tilde{y}_{ik})\}
\end{equation}
%

\iclr{

We then adopt the contrastive learning mechanism of SimSiam~\cite{simsiam} to peform the learning upon $\mathbf{V}_{\text{SDA}}$, where the loss function for each single view $z_{ik}^{(j)}=\mathcal{G}(\mathcal{F}(\tilde{x}_{ik}^{(j)}))$ is defined as follows to encourage the anchor view $\tilde{x}_{ik}^{(j)}$ being grouped up with its corresponding positive views: 
\begin{equation}
    \mathcal{L}_{\text{vw}}(z_{ik}^{(j)}) = -\sum\limits_{j' \ne j} \text{CosineSimilarity}(\mathcal{P}(z_{ik}^{(j)}), \textit{stopgrad}(z_{ik}^{(j')}))
\end{equation}
in which $\mathcal{P}$ is the predictor network and \textit{stopgrad} denotes the stop-gradient operation.
Finally, the self-supervised objective function averaged over $\mathcal{L}_{\text{vw}}$ of all views is adopted in each iteration:
\begin{equation} \label{eq:ss}
\mathcal{L}_{\text{SS}}=\sum\limits_{z_{ik}^{(j)}\in A} \mathcal{L}_{\text{vw}}(z_{ik}^{(j)})
\end{equation}
where 
$A= \{z_{ik}^{(j)}|z_{ik}^{(j)}=\mathcal{G}(\mathcal{F}(\tilde{x}_{ik}^{(j)})),~~ \forall (\tilde{x}_{ik}^{(j)},\tilde{y}_{ik}) \in \mathbf{V}_{\text{SDA}}\}$ includes the features of all views in a batch.
}

%

\iclr{
Despite the contrastive-learning-based loss $\mathcal{L}_{\text{SS}}$, we also leverage the label information by including the softmax classifier and the cross-entropy loss $\mathcal{L}_{\text{CE}}$:
\begin{equation}
    \mathcal{L}_{\text{CE}} = - \sum_i \sum_k \mathbf{1}(\tilde{y}_{ik})^T \log p_{ik}^{(j)}
\end{equation}
where given a view $\tilde{x}_{ik}^{(j)}$, $p_{ik}^{(j)}=\text{softmax} (W^T\mathcal{F}(\tilde{x}_{ik}^{(j)})+b)$ is a probability vector with length $K(C_{old}+C_t)$, $W$ is a weight matrix, $b$ is a bias vector, the one-hot vector $\mathbf{1}(\tilde{y}_{ik})$ has value $1$ for the element indexed by $\tilde{y}_{ik}$ and zero everywhere else.

}

\subsection{Weight-Aware Balanced Sampling}
\iclr{The softmax classifier is likely to be biased towards the classes with more training samples~\cite{kang2019decoupling,wu2019bic,zhao2020wa}}. In the online continual learning scenario, the learning agent accesses more samples related to new classes because only a small fraction of the old examples are stored. Thus, the model tends to classify samples into new classes. To tackle such an issue, as every column of the weight matrix $W$ for the softmax classifier represents the weights for the corresponding class (hence being related to the degree of bias),
we propose weight-aware balanced sampling (WABS) which adaptively decides the sample ratio between old and new classes to balance the classifier based on degree of bias.
We first define a sampling rate $\gamma $ as follows:
\begin{equation} \label{eq:rate}
    \gamma=\min{(1,\frac{2\times\text{exp}(w_{\text{old}}/\tau_w)}{\text{exp}(w_{\text{old}}/\tau_w)+\text{exp}(w_{\text{new}}/\tau_w)})}
\end{equation}
where $w_{\text{new}}$ is the mean over all the weights related to the new classes (i.e. average over the columns in $W$ corresponding to the new classes) and \iclr{$w_{\text{old}}$} is defined similarly for the old classes, and $\tau_w$ is a hyperparameter.

Then we reformulate the cross-entropy loss as follows:
\begin{equation}
    \mathcal{L}_{\text{WABS}} = - \sum_i \sum_k \mathbb{1}_{\text{WABS}}(\tilde{y}_{ik})\mathbf{1}(\tilde{y}_{ik})^T \log p_{ik}^{(j)}
\end{equation}
where $\mathbb{1}_{\text{WABS}}$ is defined as below, with uniformly drawing $\Gamma$ from $[0,1]$ for each sample:
\begin{equation}
    \mathbb{1}_{\text{WABS}} (\tilde{y}_{ik})= 
\begin{cases}
    1,& \text{if } \tilde{y}_{ik} \le K C_{\text{old}} \\
    1,& \text{if } \tilde{y}_{ik} > K C_{\text{old}} ,~ \Gamma < \gamma\\
    0,& \text{if } \tilde{y}_{ik} > K C_{\text{old}} ,~ \Gamma \ge \gamma\\
\end{cases}
\end{equation}
in which we keep all those views belonging to the old classes while for each of the views belonging to new classes it has $\gamma$ probability to be kept, for the use in the cross-entropy loss. Please note that, the WABS only applies on $\mathcal{L}_{\text{WABS}}$ while \iclr{$\mathcal{L}_{\text{SS}}$} uses all views without any sampling.

The overall objective $ \mathcal{L}_{\text{total}}$ for our full model (named \textbf{SDAF}) combines the proposed \iclr{$\mathcal{L}_{\text{SS}}$} to reduce forgetting by learning on diverse views and the cross-entropy loss $\mathcal{L}_{\text{WABS}}$ with adaptive sampling to \iclr{utilise the label information}:
\begin{equation}
    \mathcal{L}_{\text{total}} = \mathcal{L}_{\text{WABS}} + \lambda \mathcal{L}_{\text{SS}}
    \label{eq:total_loss}
\end{equation}
where $\lambda$ controls the balance between two losses. 
    
\subsection{Inference}\label{inference}
We adopt nearest-centre-mean (NCM) classifier for inference. For any test sample $x$, we average its distance to all class centres over $K$ augmentations. First, we calculate the centres $m_{ck}$ for all $K(C_{old}+C_t)$ classes, 
\begin{equation}
    m_{ck}=\frac{1}{|R_{ck}|} \sum_{r_{ik} \in R_{ck}} r_{ik}
\end{equation}
where $R_{ck} = \{r_{ik}=\mathcal{F}(S_k(x_i)) | y_i = c, ~ (x_i,y_i) \in \mathcal{M} \}$.
And we define the prediction function for a test sample $x$ as:
\begin{equation}
    \hat{y}=\operatorname*{argmin}_c \frac{1}{K} \sum_k d(\mathcal{F}(S_k(x)), m_{ck})
    \label{eq:prediction}
\end{equation}

The distance metric $d(x, m) = \sqrt{(x-m)^T \Sigma^{-1} (x-m)}$ is based on the Mahalanobis distance and $\Sigma^{-1} = \text{Cov}^{-1}(\mathbf{R})$ is the pseudo-inverse of the covariance matrix of the set $\mathbf{R}=\bigcup_{c \in \mathcal{C_{\text{old}}} \cup \mathcal{C}_t}\bigcup_{k=1}^K R_{ck}$. \iclr{Noting that for previous methods (e.g. SCR) they typically adopt Euclidean distance for $d(x, m)$ in the nearest-centre-mean classifier. The reason behind our using Mahalanobis distance is that it takes the feature distribution into consideration via covariance matrix, while Euclidean distance only computes the distance from every individual sample to the mean of exemplars.}

\section{Experimental Results}

\subsection{Datasets}
We conduct experiments on three popular benchmarks, including CIFAR-10 \cite{alex2009cifar}, CIFAR-100~\cite{alex2009cifar}, and mini-imagenet~\cite{deng2009imagenet,vinyals2016matching}.
We split CIFAR-10 into 5 incremental stages, and each stage contains 2 classes. We split CIFAR-100 into 10 incremental stages, and each stage contains 10 classes. We split Mini-Imagenet into 10 incremental stages and each stage contains 10 classes.

\subsection{Quantitative Metrics}
We adopt several evaluation metrics to assess the performance of online continual learning, including: average incremental accuracy ($\mathbb{A}$), end accuracy ($\mathbb{E}$), forgetting measure ($\mathbb{F}$)~\cite{chaudhry2018ewcpp}, centered kernel alignment ($\mathbb{CKA}$)~\cite{kornblith2019cka}, and balance measure ($\beta$)~\cite{kang2021exploring}, in which they are detailed in Appendix~\ref{quantitative_metric}.

\begin{table*}[tbh!]
\vspace{-1\baselineskip}
\centering
\caption{Evaluation results in terms of end accuracy ($\mathbb{E}$) $\pm$ std (average over \iclr{3} random orders of class arrival in the data stream). All methods are trained with the similar computational cost. 
}
\label{tab:results_1x}
\resizebox{\textwidth}{!}{\begin{tabular}{l|rrrrrrrrr}
\toprule
\multicolumn{1}{c}{Methods} & \multicolumn{3}{|c|}{Mini-ImageNet} & \multicolumn{3}{|c|}{CIFAR-100} & \multicolumn{3}{|c}{CIFAR-10} \\
\multicolumn{1}{c}{} & \multicolumn{1}{|c|}{$\mathcal{M}$=1000} & \multicolumn{1}{c|}{$\mathcal{M}$=2000} & \multicolumn{1}{c|}{$\mathcal{M}$=5000} & \multicolumn{1}{|c|}{$\mathcal{M}$=1000} & \multicolumn{1}{c|}{$\mathcal{M}$=2000} & \multicolumn{1}{c|}{$\mathcal{M}$=5000} & \multicolumn{1}{|c|}{$\mathcal{M}$=200} & \multicolumn{1}{c|}{$\mathcal{M}$=500} & \multicolumn{1}{c}{$\mathcal{M}$=1000}\\
\midrule
EWC++ & - & 4.5 $\pm$ 0.2 & - & - & 5.8 $\pm$ 0.3 & - & - & 18.1 $\pm$ 0.3 & -  \\
ER & 9.3 $\pm$ 0.8 & 12.1 $\pm$ 1.5 & 20.1 $\pm$ 1.8 & 11.1 $\pm$ 0.1 & 14.2 $\pm$ 0.7 & 20.6 $\pm$ 0.9 & 24.1 $\pm$ 3.0 & 29.1 $\pm$ 4.1 & 38.2 $\pm$ 3.4 \\
AGEM & 5.0 $\pm$ 0.8 & 5.1 $\pm$ 0.9 & 5.2 $\pm$ 0.6 & 6.1 $\pm$ 0.5 & 6.1 $\pm$ 0.5 & 6.1 $\pm$ 0.5 & 18.1 $\pm$ 1.4 & 18.2 $\pm$ 1.2 & 18.3 $\pm$ 0.9 \\
GSS & 8.4 $\pm$ 0.8 & 11.1 $\pm$ 2.7 & 14.9 $\pm$ 2.5 & 10.4 $\pm$ 0.4 & 12.6 $\pm$ 0.7 & 16.9 $\pm$ 1.1 & 20.3 $\pm$ 1.7 & 24.7 $\pm$ 2.9 & 32.0 $\pm$ 5.2 \\
MIR & 8.8 $\pm$ 0.5 & 10.9 $\pm$ 1.0 & 18.5 $\pm$ 1.2 & 10.9 $\pm$ 0.4 & 13.6 $\pm$ 0.7 & 19.0 $\pm$ 0.9 & 22.9 $\pm$ 3.2 & 29.6 $\pm$ 4.0 & 37.2 $\pm$ 4.2 \\
ASER & 13.7 $\pm$ 1.4 & 16.8 $\pm$ 1.7 & 24.8 $\pm$ 1.3 & 13.2 $\pm$ 0.8 & 17.3 $\pm$ 0.8 & 23.3 $\pm$ 1.0 & 22.4 $\pm$ 3.2 & 28.0 $\pm$ 4.3 & 32.5 $\pm$ 3.2 \\
DualNet & 15.8 $\pm$ 0.6 & 22.9 $\pm$ 1.3 & 27.0 $\pm$ 3.0 & 16.7 $\pm$ 2.2 & 21.5 $\pm$ 1.6 & 25.0 $\pm$ 1.6 & 44.3 $\pm$ 2.7 & 52.8 $\pm$ 2.4 & 56.0 $\pm$ 3.1 \\
DVC & 22.2 $\pm$ 0.7 & 27.4 $\pm$ 0.9 & \textbf{33.4 $\pm$ 0.5} & 25.4 $\pm$ 0.7 & 30.5 $\pm$ 0.6 & 36.6 $\pm$ 1.6 & 48.2 $\pm$ 3.0 & 55.6 $\pm$ 2.6 & 59.8 $\pm$ 4.1 \\
SCR & 15.8 $\pm$ 1.5 & 16.4 $\pm$ 2.0 & 17.7 $\pm$ 1.8 & 20.9 $\pm$ 1.2 & 22.1 $\pm$ 1.4 & 24.1 $\pm$ 0.9 & 44.6 $\pm$ 6.6 & 58.4 $\pm$ 5.1 & 65.7 $\pm$ 2.6 \\
SCL & 14.6 $\pm$ 0.8 & 15.8 $\pm$ 1.2 & 16.6 $\pm$ 1.3 & 18.8 $\pm$ 1.1 & 20.4 $\pm$ 1.3 & 22.0 $\pm$ 0.9 & 49.9 $\pm$ 5.8 & 61.0 $\pm$ 1.8 & 66.6 $\pm$ 1.5 \\
SDAF & \textbf{22.7 $\pm$ 0.5} & \textbf{28.3 $\pm$ 0.3} & 33.2 $\pm$ 0.5 & \textbf{29.3 $\pm$ 2.1} & \textbf{35.3 $\pm$ 0.7} & \textbf{39.0 $\pm$ 0.3} & \textbf{52.9 $\pm$ 3.5} & \textbf{66.4 $\pm$ 1.0} & \textbf{70.1 $\pm$ 0.5}\\
\bottomrule
\end{tabular}}
\end{table*}

\subsection{Results}
\textbf{End Accuracy \texorpdfstring{$\mathbb{E}$}{E}}.
We compare our full model (SDAF) with respect to several state-of-the-art baselines, including EWC++~\cite{chaudhry2018ewcpp}, ER~\cite{chaudhry2019er}, AGEM~\cite{AGEM}, GSS~\cite{NEURIPS2019gss}, MIR~\cite{NIPS2019_mir}, ASER~\cite{shim2021aser}, DualNet~\cite{pham2021dualnet}, \iclr{DVC~\cite{gu2022dvc}}, SCR~\cite{mai2021scr}, and SCL~\cite{khosla2020supervised}, in which the corresponding explanations of these methods have been provided in Appendix~\ref{baselines}. 
As our proposed method multiplies the size of a batch by 8 times (i.e. firstly adopting semantically distinct augmentation to augment an image into $K=4$ instances, followed by generating $2$ views for each of the instances), 
in order to have the fair comparison among our proposed method and the baselines in terms of the same computational cost, 
we set the number of SGD update $I$ to 1 for our proposed method, $I$ to 4 for the baselines based on contrastive learning \iclr{(e.g. SCR, SCL, DualNet, and DVC)}, and $I$ to 8 for the other methods.
\walon{Please note that, as DualNet contains two components (i.e. a slow learner adopting self-supervised learning and a fast learner adopting supervised learning), both of these two components will run for $I=4$ SGD updates before receiving the next new batch. We follow the aforementioned settings of $I$ for SGD updates in all of our experiments unless otherwise
specified.} 
\iclr{The results in terms of end accuracy $\mathbb{E}$ is shown in Table~\ref{tab:results_1x}, in which it is clear to observe that our proposed SDAF method outperforms almost of the baselines on different datasets with various settings of memory size $\mathcal{M}$ except being slightly worse than DVC on Mini-ImageNet with $\mathcal{M}=5000$ (noting that DVC has a specific strategy to draw the most informative samples from memory, where such strategy benefits more when memory size gets larger). 
}

\begin{wraptable}{o}{0.3\linewidth}\small
    \vspace{-2.5\baselineskip}
    \centering
    \caption{Forgetting ($\mathbb{F}$) $(\downarrow)$.}
    \begin{tabular}{l|r}
        \toprule
        Methods & $F$ \\
        \midrule
        DVC & 0.278 \\
        SCR & 0.328 \\
        SCL & 0.261 \\
        SDAF & \textbf{0.197} \\
        \bottomrule
    \end{tabular}
    \label{tab:forgetting}
    \vspace{-.5\baselineskip}
\end{wraptable}

\textbf{Catastrophic Forgetting}.
Table~\ref{tab:forgetting} shows the forgetting measure ($\mathbb{F}$) of several methods that are trained on CIFAR-10 \iclr{with $\mathcal{M} = 500$}, where our proposed SDAF is shown to effectively alleviate catastrophic forgetting. It is contributed to our using strong augmentations for generating diverse samples as novel classes, which leads to better learning the generalised representation and the model is able to handle the newly arriving classes with less adjustment.


\iclr{
\textbf{Ablation Study on SDA}.
Here we conduct an ablation study on different designs \iclrA{under similar computational cost} for our proposed SDA, where two variants are adopted in this study: ``Align'' means that these samples are treated as belonging to their original classes, and \iclrA{``Identity''} means that $\mathcal{S}$ is an identity function, while ``Proposed SDAF'' means that we treat samples generated by $\mathcal{S}$ as new classes as described in Section~\ref{sda}, As shown in Table~\ref{tab:sda_ablation}, the proposed method has both superior end accuracy $\mathbb{E}$ and average incremental accuracy $\mathbb{A}$, demonstrating the effectiveness of our SDA. 

\textbf{The Impact of Distance Metric in Prediction Function}.
As described in Section~\ref{inference} that we adopt Mahalanobis distance instead of typical Euclidean distance for $d(x,m)$ in the nearest-centre-mean classifier, here we investigate the corresponding impact of such design choice. As shown in Table~\ref{tab:distance}, the difference in terms of end accuracy $\mathbb{E}$ between adopting the Euclidean distance (identity matrix for $\Sigma$ in $d(x,m)$) or the Mahalanobis distance in the prediction function (cf. Eq.~\ref{eq:prediction}) increases along with the growing memory size $\mathcal{M}$ on all the benchmarks. For example, the difference in $\mathbb{E}$ between Euclidean and Mahalanobis distance are $-2.6\%, 0.3\%$, and $1.7\%$ corresponding to $\mathcal{M}$ set to $200$, $500$, and $1000$ respectively on CIFAR-10. The potential cause contributes to the fact that the covariance estimation gets more accurate when $\mathcal{M}$ increases, in the high-dimensional feature space.
}

\begin{table}[!htbp]\small
\vspace{-2\baselineskip}
\begin{minipage}[t]{.45\textwidth}
\centering
\caption{\iclr{Ablation study on design for our semantically distinct augmentation $\mathcal{S}$ (experiments based on CIFAR-10 with $\mathcal{M} = 500$).}}
\begin{tabular}{l|c|cc}
\toprule
& $I$ & $\mathbb{E}$ & $\mathbb{A}$\\
\midrule
Proposed SDAF & 1 & \textbf{66.4} & \textbf{78.5} \\
Align & 1 & 64.6 & 77.6 \\
Identity & 4 & 61.0 & 76.1\\
\bottomrule
\end{tabular}
\label{tab:sda_ablation}
\end{minipage}
\hfill
\begin{minipage}[t]{.45\textwidth}
\centering
\caption{\iclr{The comparison of end accuracy ($\mathbb{E}$) on adopting Euclidean or Mahalanobis distance in the prediction function (cf. Eq.~\ref{eq:prediction}). This experiment conducts on CIFAR-10.}}
\begin{tabular}{llll}
\toprule
Memory size ($\mathcal{M}$) & 0.2k & 0.5k & 1.0k \\
\midrule
Euclidean                   & \textbf{55.5}   & 66.1   & 68.4   \\
Mahalanobis                 & 52.9   & \textbf{66.4}   & \textbf{70.1}   \\
\bottomrule
\end{tabular}
\label{tab:distance}
\end{minipage}
\end{table}
\vspace{-1\baselineskip}

\iclr{Please refer to our Appendix for discussion on baselines, \textbf{more results}, and implementation details.}

\section{Conclusion}
\iclr{We propose a class-incremental online continual learning approach which stems from the basic idea of learning generalised representation to alleviate the issue of catastrophic forgetting, where we particularly utilise the strong augmentation for producing the semantically distinct classes to enhance the generalised representation learning as well as additionally adopt the softmax classifier and the weight-aware balanced sampling strategy to tackle the imbalanced dataset. With a similar computational cost, our method provides superior performance with respect to several baselines.}


\bibliographystyle{plain}
\bibliography{reference}

\clearpage
\appendix
\section{Appendix}

\walon{
\subsection{Baselines} \label{baselines}
We compare our proposed SDAF with a suite of state-of-the-art online continual learning algorithms:
First, EWC++~\cite{chaudhry2018ewcpp} is an online version of elastic weight consolidation (EWC)~\cite{kirkpatrick2017ewc} that regularises the update of the crucial weights for the past classes, in which no replay buffer is used in such approach;
Experience replay (ER)~\cite{chaudhry2019er} and all the following baselines are built upon a replay buffer $\mathcal{M}$, where ER is among the simplest one which adopts reservior sampling~\cite{10.1145/3147.3165reservoir} for MemoryUpdate operations and uniform sampling for MemoryRetrieval as introduced in Algorithm \ref{alg:algorithm} (noting that all the baselines described below also adopt the same sampling strategies for the MemoryUpdate and MemoryRetrieval operations unless otherwise specified);
Averaged gradient episodic memory (AGEM)~\cite{AGEM} utilises the samples in the replay buffer to constrain the update of the parameters;
%
Gradient-based sample selection (GSS)~\cite{NEURIPS2019gss} is a replay update method that improves MemoryUpdate operation by selecting samples with diversification of gradients to be stored in the replay memory; 
Maximally interfered retrieval (MIR)~\cite{NIPS2019_mir} provides an improved replay retrieval algorithm (related to MemoryRetrieval) that finds the samples impacted by the current batch for performing training iterations; 
Adversarial shapley value experience replay (ASER)~\cite{shim2021aser} leverages Shapley value to determine the contribution of memory samples for learning performance, and develop the corresponding strategies to improve both the memory update and retrieval operations;
DualNet~\cite{pham2021dualnet} contains two components, called fast learner and slow learner. When it gets a new batch of data $B^u_t$, it first updates the slow learner by self-supervised learning (i.e. Barlow twins~\cite{zbontar2021barlow}) on the samples from memory ($B_M$), then it performs supervised learning to update both the fast learner and slow learner on $B^u_t \bigcup B_M$; 
Supervised contrastive replay (SCR)~\cite{mai2021scr} leverages the supervised contrastive learning to learn the representation for better continual learning performance, and it observed that the nearest class mean (NCM) classifier is better to deal with the recency bias than the softmax classifier;
SCL~\cite{khosla2020supervised} is not an online continual learning method initially. We borrow it from representation learning, with having all the settings to follow the SCR ones except the transformation module $\mathcal{H}$ as described in Figure \ref{fig:clr}.
%
}
\iclr{DVC~\cite{gu2022dvc} proposes the dual view consistency (which highly relates to contrastive learning but instead adopts mutual information in its objective) to fully utilises the single-pass data stream. Moreover, DVC provides an improved replay retrieval algorithm (related to MemoryRetrieval, cf. Algorithm~\ref{alg:algorithm} in the main manuscript) that selects the samples whose gradients generated in the network are most interfered by the new incoming samples, but such retrieval algorithm actually consumes 10 times forward operations (where the cost is estimated according to its hyperparameter setting) to select a single sample for replay, which is extremely costly for OCL thus being relatively impractical.} 

\walon{
\subsection{Quantitative Metrics} \label{quantitative_metric}
\textbf{Average Incremental Accuracy} ($\mathbb{A}$) measures the accuracy over all the stages. 
Given that the learning target of the recognition system at stage $t$ is to achieve a $\mathbf{C}^t$-ways classification (i.e. there are in total $\mathbf{C}^t$ classes till stage $t$), and we denote $a_{t,c}$ the recognition accuracy for a class $c \in \mathbf{C}^t$ at the stage $t$ where $t=1,2,\dots,T$ and $c=1,2,\dots,\mathbf{C}^t$, the average incremental accuracy $\mathbb{A}$ is then defined as  $\mathbb{A}=\frac{1}{T}\sum_{t=1}^{T}\frac{1}{\mathbf{C}^t}\sum_{c=1}^{\mathbf{C}^t} a_{t,c}$\\
%
%
\textbf{End Accuracy} ($\mathbb{E}$) is the accuracy of the last stage $T$ which averages over $\mathbf{C}^T$ classes. That is, $\mathbb{E}=\frac{1}{\mathbf{C}^T}\sum_{c=1}^{\mathbf{C}^T} a_{T,c}$\\
\textbf{Forgetting Measure} ($\mathbb{F}$) is the average deterioration of accuracy over all the classes, and lower forgetting measure ($\mathbb{F}$) implies less forgetting on the previously learnt classes. 
With denoting $f_{t,c}$ the deterioration on accuracy for a class $c$ (i.e. the maximum decrease on accuracy for classifying class $c$ since the first time that the model learns to recognize it at stage $t'$) after the model finishes the training stage $t$, i.e.
\begin{equation}
    f_{t,c}=\max_{l\in \{t',\dots,t-1\}} a_{l,c} - a_{t,c},
\end{equation}
the forgetting measure $\mathbb{F}$ is then defined as following:
\begin{equation}
    \mathbb{F} = \frac{1}{T-1} \sum_{t=2}^T \frac{1}{C^{t-1}} \sum_{c=1}^{C^{t-1}} f_{t,c}
\end{equation}
%
\textbf{Centred Kernel Alignment} ($\mathbb{CKA}$) proposed by~\cite{kornblith2019cka} is a metric to measure the representation similarity between two networks.
Here in this work, we leverage the linear CKA (as later shown in Eq.~\ref{eq:cka}) to evaluate the similarity between the feature representations (obtained by the feature extractor of our model) learnt by our models in different stages, in order to observe how much the network is changed/updated across stages. We denote the feature extractor in our learnt model after finishing the training stage $t$ as $\mathcal{F}_t$, the data available or used in the training stage $\dot{t}$ as $D_{\dot{t}}$, and the collection of feature representations of $D_{\dot{t}}$ projected by $\mathcal{F}_t$ as $R_{t,\dot{t}} \in \mathbb{R}^{n \times d}$ where $n$ and $d$ denote the number of data samples in $D_{\dot{t}}$ and the feature dimensionality respectively. Given the data samples $D_{\dot{t}}$, the representation similarity between two stages $t$ and $t+1$ is now formulated as \begin{equation} \label{eq:cka}
    \mathbb{CKA}(R_{t,\dot{t}}, R_{t+1,\dot{t}})=\frac{\|R_{t,\dot{t}}^\top R_{t+1,\dot{t}}\|^2_\text{F}}{\|R_{t,\dot{t}}^\top R_{t,\dot{t}}\|_\text{F} \|R_{t+1,\dot{t}}^\top R_{t+1,\dot{t}}\|_\text{F}}
\end{equation}
where $\|\cdot\|_\text{F}$ is the Frobenius norm of a matrix, with noting that $R_{\cdot,\cdot}$ is already centered (i.e. subtracting the mean) in this computation.


A larger $\mathbb{CKA}$ value implies a higher similarity between two \iclrA{representations or models; Hence, less forgetting between stages}. Moreover, $\mathbb{CKA}$ is invariant to orthogonal transformation of representations, and the normalisation term of $\mathbb{CKA}$ ensures that it is invariant to scaling. \\
%
\textbf{Balancedness} ($\beta$) \cite{kang2021exploring} measures the classification uniformity of a classifier, in which we adopt it to examine whether the model is biased towards new classes or not at the last stage. The balancedness ($\beta$) is defined as following:
\begin{equation}
    \beta=\frac{1}{(C^T)^2}\sum_{i,j=1}^{C^T}\exp \left( -\frac{|a_{T,i}-a_{T,j}|^2}{\sigma} \right)
\end{equation}
where the scaling parameter $\sigma$ is set to $0.5$. The metric achieves maximum when the accuracy values for all classes are equal, indicating that the model does not have any bias.
}

\subsection{More Results}

\subsubsection{Average Incremental Accuracy \texorpdfstring{$\mathbb{A}$}{A}}
\iclr{The results in terms of average incremental accuracy $\mathbb{A}$ is shown in Table \ref{tab:results_1x_avg_incremental_accuracy}, in which it is clear to observe that our proposed SDAF method outperforms almost of the baselines on different datasets with various settings of memory size $\mathcal{M}$.
}

\begin{table*}[ht]\small
\centering
\caption{Evaluation results in terms of average incremental accuracy ($\mathbb{A}$) $\pm$ std (average over \iclr{3} random orders of class arrival in the datastream). All methods are trained with the similar computational cost. 
}
\resizebox{\textwidth}{!}{\begin{tabular}{l|rrrrrrrrr}
\toprule
\multicolumn{1}{c}{Methods} & \multicolumn{3}{|c|}{Mini-ImageNet} & \multicolumn{3}{|c|}{CIFAR-100} & \multicolumn{3}{|c}{CIFAR-10} \\
\multicolumn{1}{c}{} & \multicolumn{1}{|c|}{M=1k} & \multicolumn{1}{c|}{M=2k} & \multicolumn{1}{c|}{M=5k} & \multicolumn{1}{|c|}{M=1k} & \multicolumn{1}{c|}{M=2k} & \multicolumn{1}{c|}{M=5k} & \multicolumn{1}{|c|}{M=0.2k} & \multicolumn{1}{c|}{M=0.5k} & \multicolumn{1}{c}{M=1k}\\
\midrule
EWC++ & - & 6.4 $\pm$ 0.8 & - & - & 8.9 $\pm$ 1.1 & - & - & 35.8 $\pm$ 4.6 & - \\
ER & 21.6 $\pm$ 2.4 & 25.2 $\pm$ 3.0 & 32.2 $\pm$ 2.9 & 24.8 $\pm$ 1.3 & 29.3 $\pm$ 1.6 & 34.5 $\pm$ 1.8 & 48.2 $\pm$ 2.0 & 53.7 $\pm$ 2.8 & 59.7 $\pm$ 3.1\\
AGEM & 10.4 $\pm$ 1.2 & 10.5 $\pm$ 1.0 & 10.6 $\pm$ 0.9 & 14.9 $\pm$ 0.6 & 15.0 $\pm$ 0.5 & 14.9 $\pm$ 0.5 & 40.8 $\pm$ 1.3 & 40.8 $\pm$ 1.3 & 40.8 $\pm$ 1.4\\
GSS & 20.3 $\pm$ 3.2 & 23.7 $\pm$ 3.4 & 27.4 $\pm$ 3.4 & 24.3 $\pm$ 1.7 & 27.1 $\pm$ 2.0 & 31.0 $\pm$ 2.1 & 43.4 $\pm$ 0.9 & 46.5 $\pm$ 0.8 & 53.9 $\pm$ 2.9\\
MIR & 20.9 $\pm$ 2.1 & 24.0 $\pm$ 1.4 & 31.1 $\pm$ 2.9 & 24.5 $\pm$ 1.6 & 28.1 $\pm$ 1.7 & 33.6 $\pm$ 2.5 & 46.9 $\pm$ 2.0 & 53.6 $\pm$ 2.0 & 58.4 $\pm$ 2.3\\
ASER & 24.8 $\pm$ 2.7 & 29.5 $\pm$ 2.0 & 35.9 $\pm$ 2.1 & 26.2 $\pm$ 0.6 & 30.8 $\pm$ 1.3 & 36.4 $\pm$ 1.9 & 44.6 $\pm$ 1.4 & 48.3 $\pm$ 1.4 & 53.0 $\pm$ 2.6\\
DualNet & 27.2 $\pm$ 1.5 & 30.5 $\pm$ 2.5 & 32.6 $\pm$ 1.4 & 28.9 $\pm$ 2.0 & 31.9 $\pm$ 2.4 & 33.7 $\pm$ 2.3 & 61.3 $\pm$ 0.9 & 65.2 $\pm$ 2.4& 67.4 $\pm$ 2.0\\
DVC & 33.4 $\pm$ 1.0 & 37.0 $\pm$ 0.8 & \textbf{41.5 $\pm$ 0.9} & 36.9 $\pm$ 0.6 & 40.4 $\pm$ 1.0 & 43.8 $\pm$ 0.6 & 64.1 $\pm$ 1.6 & 67.1 $\pm$ 2.2 & 70.4 $\pm$ 2.3\\
SCR & 25.2 $\pm$ 2.6 & 25.9 $\pm$ 2.4 & 26.8 $\pm$ 2.4 & 31.2 $\pm$ 3.7 & 32.5 $\pm$ 3.5 & 33.2 $\pm$ 3.6 & 65.9 $\pm$ 3.6 & 73.6 $\pm$ 2.1 & 78.4 $\pm$ 1.2 \\
SCL & 23.3 $\pm$ 2.4 & 23.8 $\pm$ 2.0 & 25.0 $\pm$ 2.4 & 28.7 $\pm$ 3.5 & 29.2 $\pm$ 3.6 & 30.6 $\pm$ 3.1 & 68.7 $\pm$ 2.6 & 75.5 $\pm$ 1.6 & 78.2 $\pm$ 0.7 \\
SDAF & \textbf{34.2 $\pm$ 2.0} & \textbf{38.8 $\pm$ 2.0} & 41.3 $\pm$ 1.5 & \textbf{40.2 $\pm$ 2.1} & \textbf{44.4 $\pm$ 1.3} & \textbf{46.6 $\pm$ 1.8} & \textbf{71.3 $\pm$ 1.2} & \textbf{78.5 $\pm$ 0.8} & \textbf{80.7 $\pm$ 0.6}\\
\bottomrule
\end{tabular}}
\label{tab:results_1x_avg_incremental_accuracy}
\end{table*}

\iclrA{
\subsubsection{Stage-Wise Accuracy} \label{stagewise_acc}
Here in Table~\ref{tab:stagewise_acc} we provide detailed stage-wise accuracy for SCR, SCL, and our proposed SDAF, where the experiments are based on CIFAR-10 dataset with $\mathcal{M}=500$. Noting that SCR, SCL, and our SDAF are under the similar computational cost by setting $I=4$ to SCR/SCL and $I=1$ to our SDAF.
From the results, we can observe that our full model SDAF has superior accuracy on every stage as well as better average incremental accuracy $\mathbb{A}$.

\begin{table}[!htbp]\small
    \centering
    \caption{The stage-wise accuracy based on CIFAR-10 dataset with $\mathcal{M}=500$ to demonstrate the contributions of our semantically distinct augmentation (SDA).
    }
    \begin{tabular}{c | ccc}
    \toprule
    Stage & SCR & SCL & SDAF\\
    \midrule
    1 & 95.2 & 94.6 & \textbf{95.8}\\
    2 & 82.1 & 82.8 & \textbf{83.7}\\
    3 & 71.6 & 73.7 & \textbf{76.8}\\
    4 & 65.3 & 67.9 & \textbf{69.7}\\
    5 & 59.6 & 61.2 & \textbf{66.4}\\
    \midrule
    $\mathbb{A}$ & 74.8 & 76.1 & \textbf{78.5} \\
    \bottomrule
    \end{tabular}
\label{tab:stagewise_acc}
\end{table} 
}

\subsubsection{Representation Similarity}
\walon{
We use the linear centered kernel alignment (linear $\mathbb{CKA}$)~\cite{kornblith2019cka}
to measure the representation similarity between the models of different stages on CIFAR-10 (i.e. how much the model is updated/adjusted across stages), and the results are shown in Table~\ref{tab:cka} and Table~\ref{tab:cka_iter1_detailed}.
In Table~\ref{tab:cka_iter1_detailed}, the grey cells indicate that the given dataset $\mathcal{D}_{\dot{t}}$ is seen (trained) at both stages on comparison (i.e. $\mathcal{D}_{\dot{t}}$ has been used to train $\mathcal{F}_t$ and $\mathcal{F}_{t+1}$, where $\dot{t} \leq t$), and the white cells indicate that the given dataset $\mathcal{D}_{\dot{t}}$ is unseen at either one or both of the stages on comparison.
The column ``Seen'' in the Table~\ref{tab:cka} is the average linear $\mathbb{CKA}$ over the seen classes (i.e. related to grey cells in Table~\ref{tab:cka_iter1_detailed}) and the  column ``Unseen'' is the unseen classes (i.e. related to white cells in Table~\ref{tab:cka_iter1_detailed}).
Higher linear $\mathbb{CKA}$ (what is achieved by our SDAF) means that the model changes less while learning new classes (thanks to the more generalised representation); hence, it has less forgetting.}

\begin{table}[!htbp]\small
        \centering

        \centering
        \caption{Linear $\mathbb{CKA}$ $(\uparrow)$.}
        \begin{tabular}{l|rrr}
            \toprule
            Methods & Seen & Unseen\\
            \midrule
            SCR~\cite{mai2021scr} & 0.799 & 0.442 \\
            SCL~\cite{khosla2020supervised} & 0.773 & 0.456 \\
            SDAF & \textbf{0.847} & \textbf{0.644} \\
            \bottomrule
        \end{tabular}
        \label{tab:cka}
\end{table}

\begin{table}[!htbp]\small
\centering
\caption{Linear $\mathbb{CKA}$ $(\uparrow)$ of a given dataset $\mathcal{D}_{\dot{t}}$ between two feature extractors of contiguous stages, $\mathcal{F}_t$ and $\mathcal{F}_{t+1}$.}


\begin{tabular}{c|rrrrr}
\toprule
\multicolumn{1}{l|}{(a) SCR~\cite{mai2021scr}} & \multicolumn{1}{c}{$\mathcal{D}_1$} & \multicolumn{1}{c}{$\mathcal{D}_2$} & \multicolumn{1}{c}{$\mathcal{D}_3$} & \multicolumn{1}{c}{$\mathcal{D}_4$} & \multicolumn{1}{c}{$\mathcal{D}_5$}\\
\midrule
stage 1 \& 2 & \cellcolor[gray]{0.9}\textbf{0.775} & 0.442 & 0.491 & 0.254 & 0.486 \\
stage 2 \& 3 & \cellcolor[gray]{0.9}\textbf{0.799} & \cellcolor[gray]{0.9}0.816 & 0.58 & 0.358 & 0.451 \\
stage 3 \& 4 & \cellcolor[gray]{0.9}0.755 & \cellcolor[gray]{0.9}0.805 & \cellcolor[gray]{0.9}0.847 & 0.248 & 0.491 \\
stage 4 \& 5 & \cellcolor[gray]{0.9}0.759 & \cellcolor[gray]{0.9}0.795 & \cellcolor[gray]{0.9}0.88 & \cellcolor[gray]{0.9}0.759 & 0.616 \\
\midrule
\midrule
\multicolumn{1}{l|}{(b) SCL~\cite{khosla2020supervised}} & \multicolumn{1}{c}{$\mathcal{D}_1$} & \multicolumn{1}{c}{$\mathcal{D}_2$} & \multicolumn{1}{c}{$\mathcal{D}_3$} & \multicolumn{1}{c}{$\mathcal{D}_4$} & \multicolumn{1}{c}{$\mathcal{D}_5$} \\
\midrule
stage 1 \& 2 & \cellcolor[gray]{0.9}0.767 & 0.532 & 0.464 & 0.232 & 0.396 \\
stage 2 \& 3 & \cellcolor[gray]{0.9}0.782 & \cellcolor[gray]{0.9}\textbf{0.856} & 0.689 & 0.399 & 0.449 \\
stage 3 \& 4 & \cellcolor[gray]{0.9}0.660 & \cellcolor[gray]{0.9}0.760 & \cellcolor[gray]{0.9}0.87 & 0.394 & 0.38 \\
stage 4 \& 5 & \cellcolor[gray]{0.9}0.658 & \cellcolor[gray]{0.9}0.811 & \cellcolor[gray]{0.9}0.881 & \cellcolor[gray]{0.9}0.685 & 0.626 \\
\midrule
\midrule
\multicolumn{1}{l|}{(c) SDAF} & \multicolumn{1}{c}{$\mathcal{D}_1$} & \multicolumn{1}{c}{$\mathcal{D}_2$} & \multicolumn{1}{c}{$\mathcal{D}_3$} & \multicolumn{1}{c}{$\mathcal{D}_4$} & \multicolumn{1}{c}{$\mathcal{D}_5$} \\
\midrule
stage 1 \& 2 & \cellcolor[gray]{0.9}0.685 & \textbf{0.612} & \textbf{0.581} & \textbf{0.331} & \textbf{0.617} \\
stage 2 \& 3 & \cellcolor[gray]{0.9}0.787 & \cellcolor[gray]{0.9}0.838 & \textbf{0.784} & \textbf{0.629} & \textbf{0.692} \\
stage 3 \& 4 & \cellcolor[gray]{0.9}\textbf{0.867} & \cellcolor[gray]{0.9}\textbf{0.892} & \cellcolor[gray]{0.9}\textbf{0.918} & \textbf{0.688} & \textbf{0.785} \\
stage 4 \& 5 & \cellcolor[gray]{0.9}\textbf{0.843} & \cellcolor[gray]{0.9}\textbf{0.902} & \cellcolor[gray]{0.9}\textbf{0.925} & \cellcolor[gray]{0.9}\textbf{0.812} & \textbf{0.762} \\
\bottomrule
\end{tabular}

\label{tab:cka_iter1_detailed}
\end{table}

\begin{figure*}[!hbt]
     \centering
     \subfloat[SCR]{\includegraphics[width=1.5in]{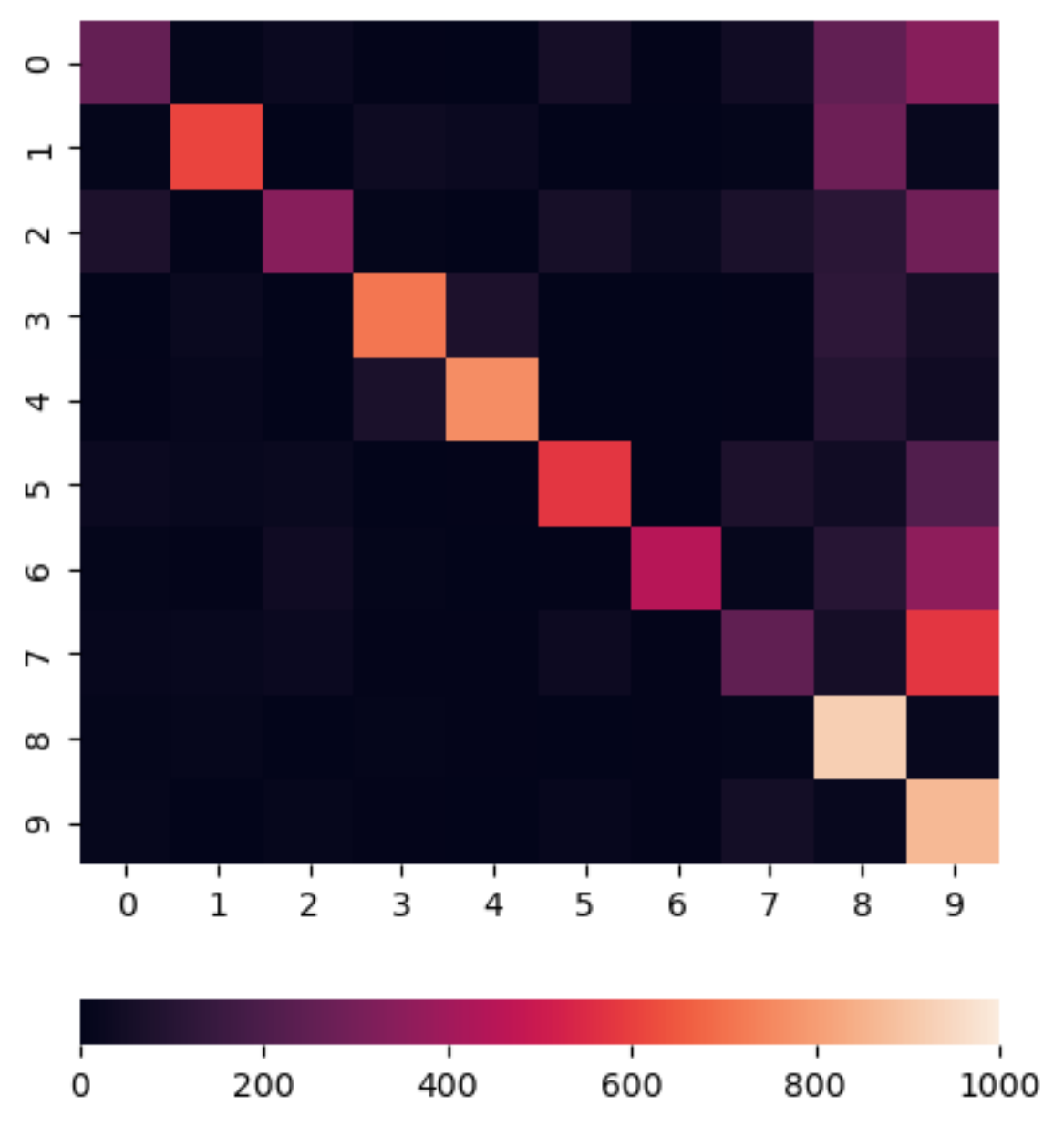}}
     \hfil
     \subfloat[SCL]{\includegraphics[width=1.5in]{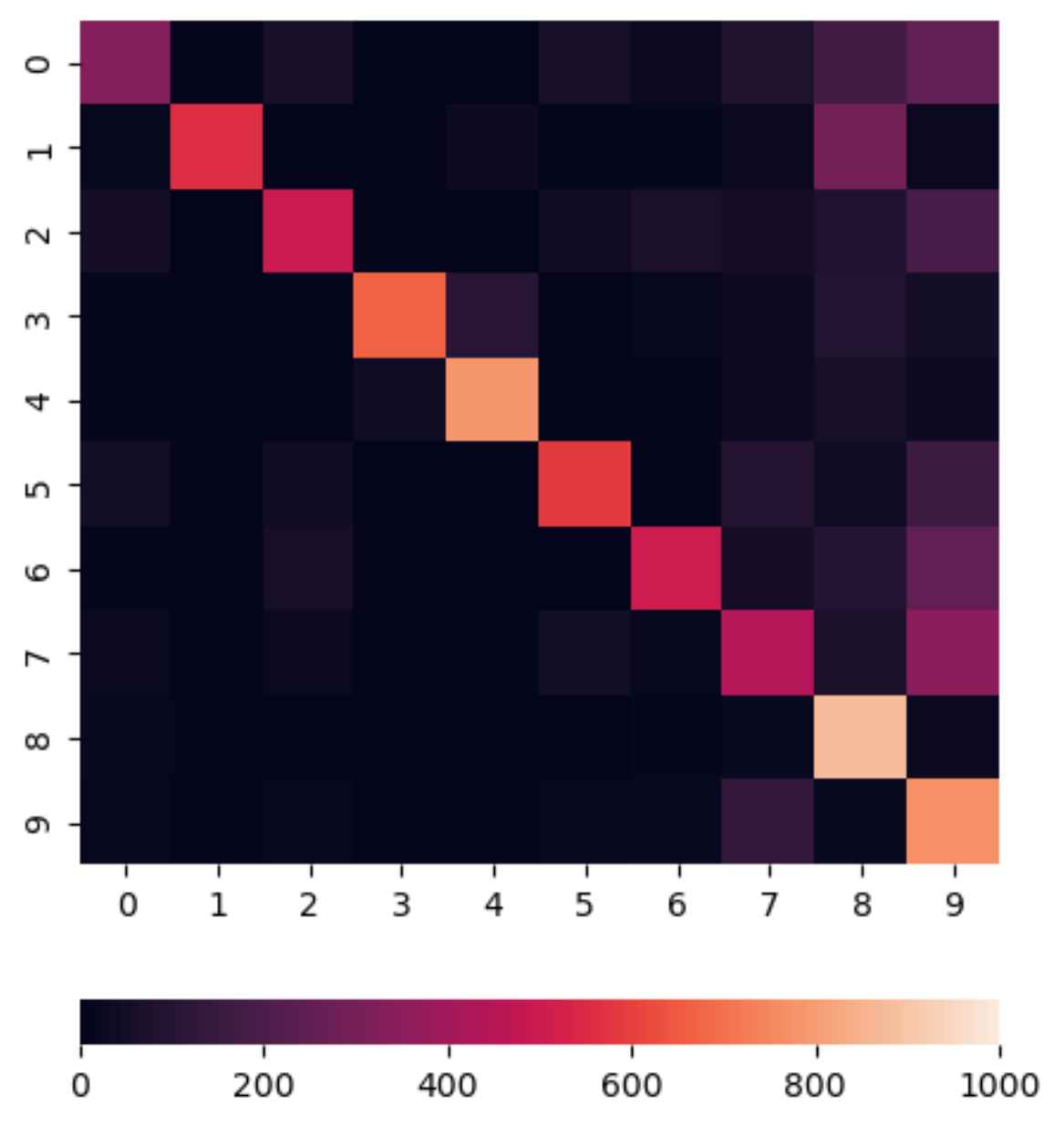}}
     \hfil
     \subfloat[SDAF]{\includegraphics[width=1.5in]{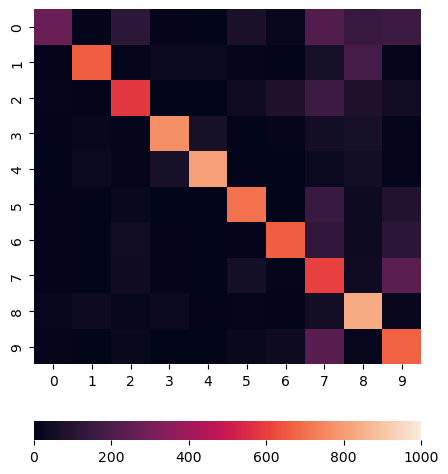}}
      \caption{The confusion matrix at 5th stage of CIFAR-10 with $\mathcal{M}$ set to 500 (noting that here we report the results of a single-run experiment instead of having the average over 5 random orders of class arrival, for the purpose of better visualising the confusion matrix). The old classes include classes 0 to 7, and the new classes includes class 8 and 9. (a) SCR. $\beta=0.737$ (b) SCL. $\beta=0.858$ (c) SDAF. $\beta=0.867$
      }
     \label{fig:confusion_matrix}

\end{figure*}

\walon{
\subsubsection{Confusion Matrix -- Balancedness \texorpdfstring{$\beta$}{beta}}
In Figure~\ref{fig:confusion_matrix}, we compare the confusion matrix of SCR, SCL, and our SDAF in a single-run experiment. All models are trained on CIFAR-10 and $\mathcal{M}$ is 500, where Figure~\ref{fig:confusion_matrix} shows the inference result of the CIFAR-10 test split using the model that completes the training of five stages. 
Our proposed SDAF model has the highest balancedness $\beta$ among three methods, in which such result together with the visualisation of confusion matrix indicate that our SDAF is least biased toward new classes (i.e. class 8 and class 9) and suffers less from an unbalanced training dataset.
}

\subsubsection{Out-of-Distribution Detection}
One way to evaluate the generalisability of a learnt representation is to testify its performance on the out-of-distribution (OOD) detection task (i.e. how likely we are able to detect the unknown samples from novel classes based on the feature representation being used). The basic motivation behind such experiments is that: Since the continual learning agent aims to sequentially extend its recognition ability on the unseen classes (i.e. unseen classes with respect to the current model, which can also be treated as out-of-distribution samples), the generalisable feature representation, which ideally would make the unseen classes more separable from the seen classes thus requires less adjustment while learning the unseen/new classes, hence is able to alleviate the catastrophic forgetting phenomenon in continual learning as the feature extractor of the classification model only needs to be minorly updated. 
To this end, we take the training set of CIFAR-10 as inliers and treat the other datasets (e.g. MNIST~\cite{lecun2010mnist}, Omniglot~\cite{lake2011one}, and SVHN~\cite{netzer2011reading}) as outliers, ideally the test samples from the test set of CIFAR-10 should be closer to the inliers than the samples from the outlier dataset (where such distance is computed via the Mahalanobis distance from the test sample to the nearest centre of inlier classes, averaged over various augmentations of the test sample). The performance of OOD detection is then evaluated by the area under ROC curve (AUC) based on the aforementioned distance, and the results are shown in Table~\ref{tab:ood}. We can see that the representation learnt by our proposed SDAF provides a better OOD detection performance than SCL and SCR thus being more generalised (with following a clear logic behind: improvement in the contrastive learning leads to better generalizability as shown by better OOD detection performance, and the better generalizability leads to better alleviation of the catastrophic forgetting thus boosting the continual learning performance).

\begin{table}[!htbp]\small
\centering
\caption{Out-of-distribution (OOD) detection result.}
\begin{tabular}{l|cccc}
\toprule
AUC & MNIST & Omniglot & SVHN \\
\midrule
SCR~\cite{mai2021scr} & 92.6 & 57.0 & 68.3 \\
SCL~\cite{khosla2020supervised} & 94.0 & 34.7 & 57.2 \\
SDAF & \textbf{95.7} & \textbf{83.7} & \textbf{88.7} \\  
\bottomrule
\end{tabular}
\label{tab:ood}
\end{table}

\begin{figure}[!htbp]
    \centering
    \vspace{-0.25in}
    \subfloat[SCR]{\includegraphics[scale=0.25]{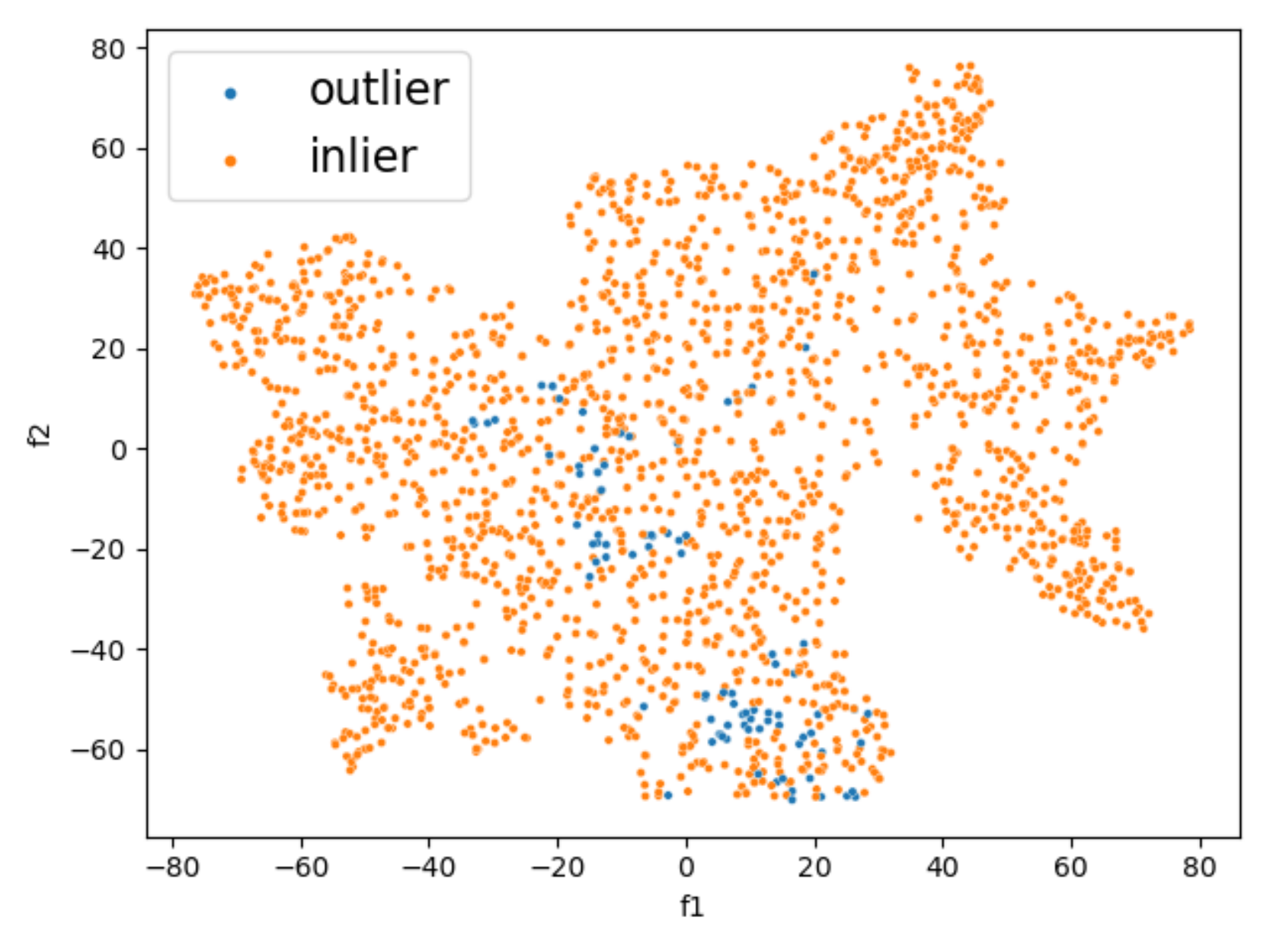}\label{fig:badtsne}}
    \hfil
    \subfloat[SDAF]{\includegraphics[scale=0.25]{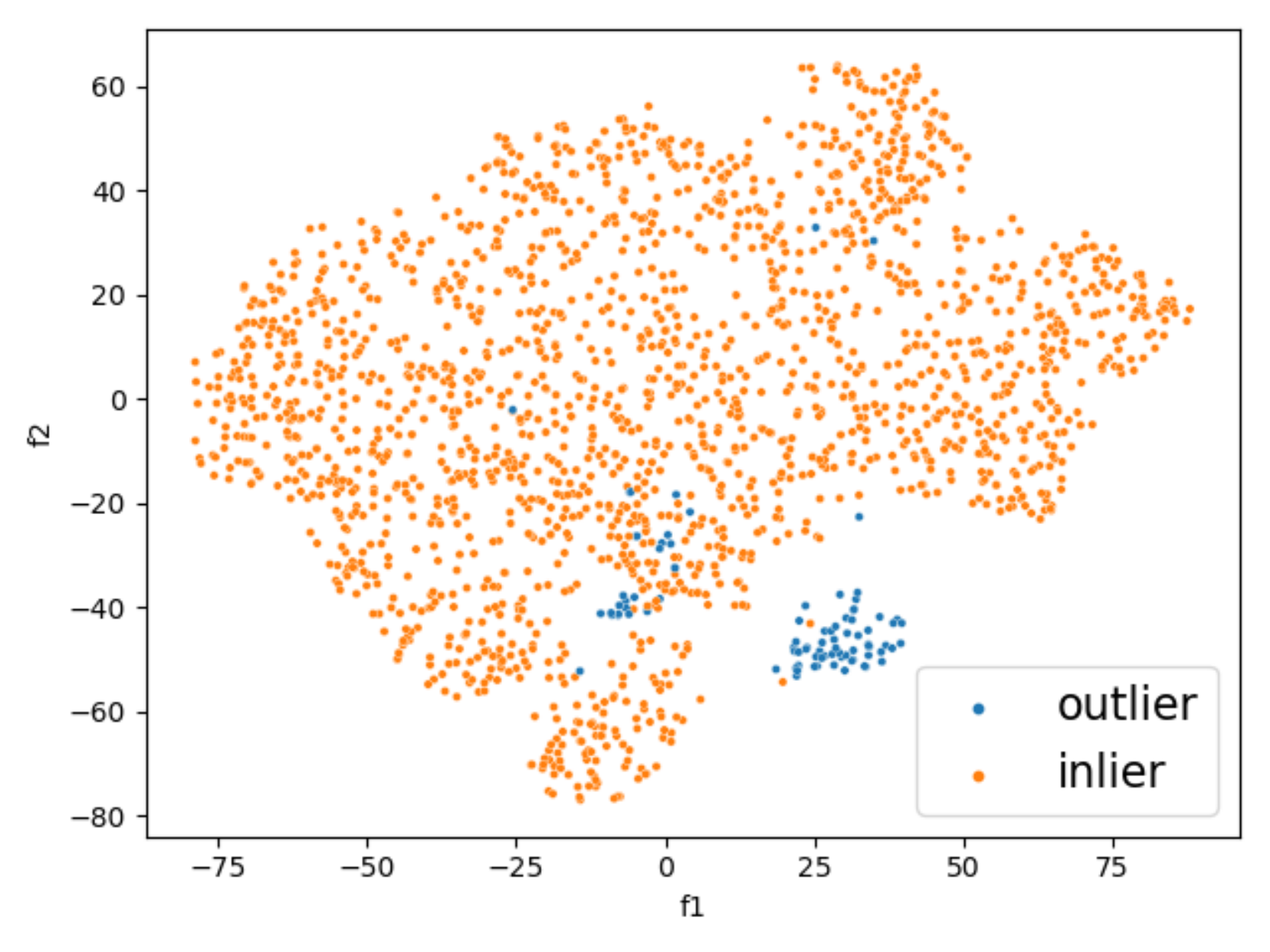}\label{fig:goodtsne}}
    \caption{tSNE visualisation of the features learnt by SCR and our SDAF. Both SCR and our SDAF are trained till the end of 5th stage of CIFAR-10. Inlier points (colorized by orange) are from the test split of CIFAR-10, and the outlier points (colorized by blue) are from the test split of the SVHN dataset. We are able to see that our SDAF better separates the inlier and outlier than what SCR can do.}
    \label{fig:tsne}

\end{figure}

\walon{
\subsubsection{Representation Visualisation}
Figure \ref{fig:tsne} is the tSNE~\cite{tSNE} visualisation of the features learnt by SCR and our SDAF, both models complete the training of the 5th stages on CIFAR-10 dataset with $\mathcal{M}$ being set to 500. This visualisation result shows the effectiveness of our SDAF method in separating trained classes (i.e. inlier, from the test split of CIFAR-10 dataset) from unseen classes (i.e. outlier, from the test split of SVHN dataset), which actually implies that our SDAF method is better generalised to unseen classes hence will suffer less from the forgetting issue while learning to recognize the new classes.
}

\begin{figure}[!htb]
\centering
\begin{minipage}[t]{.4\textwidth}
    \centering
    \includegraphics[scale=0.30]{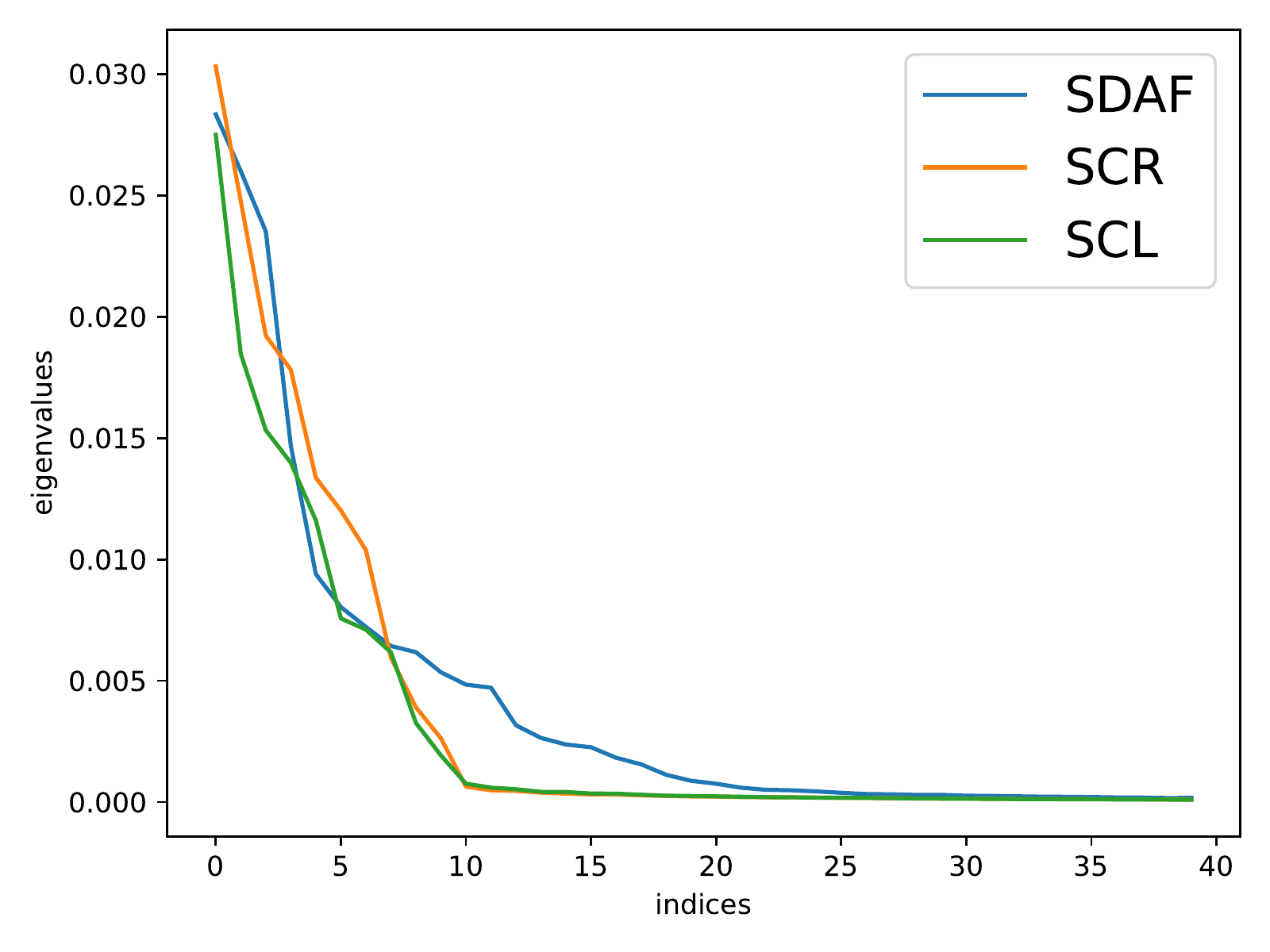}
    \caption{Scree plot of normalised representations on CIFAR-10 dataset with $\mathcal{M}$ set to 500.}
    \label{fig:scree}
\end{minipage}%
\hfil
\begin{minipage}[t]{.4\textwidth}
    \centering
    \includegraphics[scale=0.35]{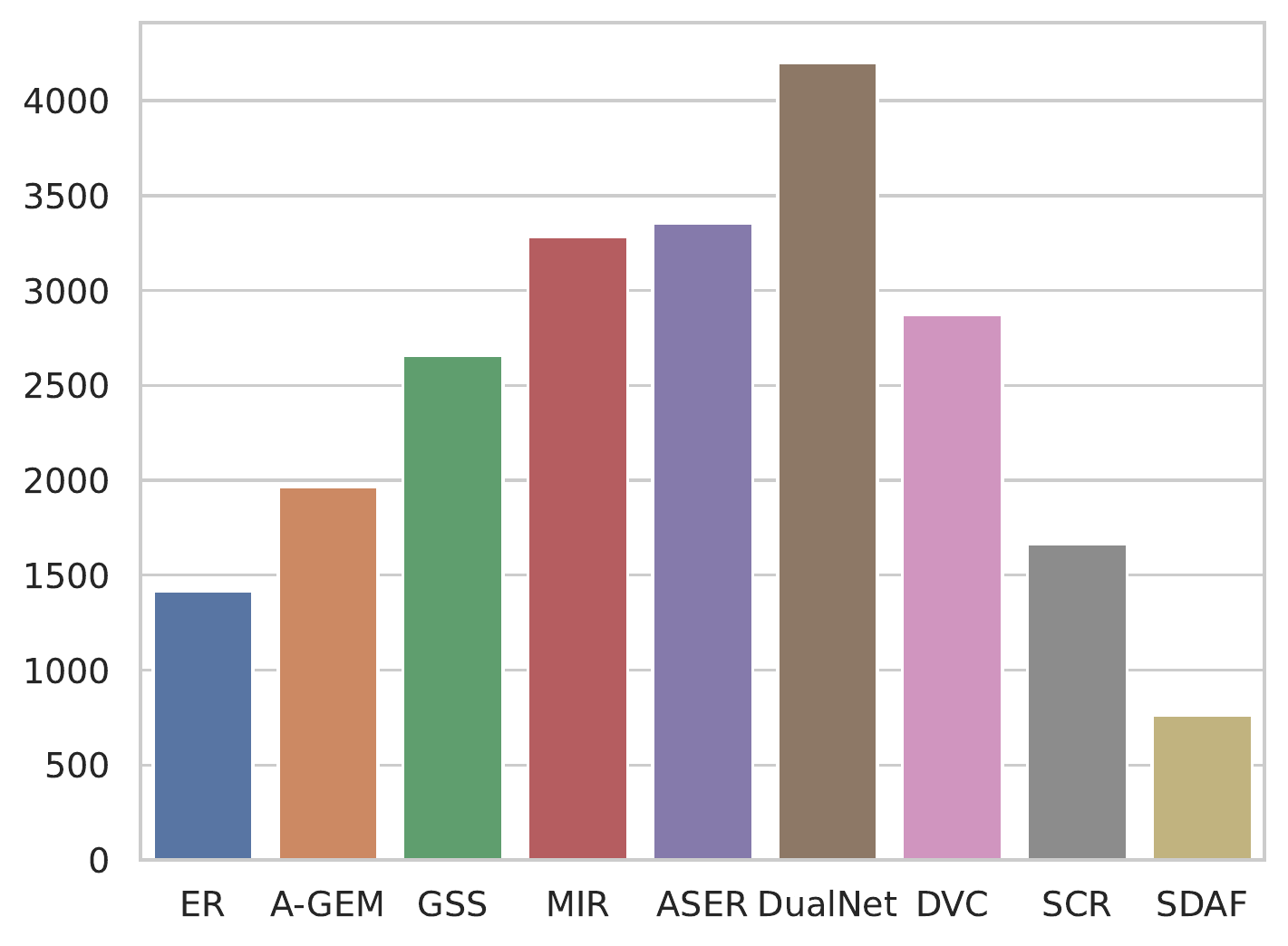}
    \caption{Executive time (seconds) of 5 incremental stages on CIFAR-10 with $\mathcal{M}$ set to 500.}
    \label{fig:time}
\end{minipage}
\end{figure}

\walon{
\subsubsection{Representation Diversity}
Figure~\ref{fig:scree} shows the scree plot (i.e. the line plot of the eigenvalues) for the covariance matrix of the normalised feature representations (with the maximum norm being one) obtained from various methods (i.e. our SDAF, SCL, and SCL) which are trained on CIFAR-10 with $\mathcal{M}$ set to 500. 
The $x$-axis of the scree plot presents the indices of the eigenvalues which are sorted in a descending order, and the $y$-axis presents the corresponding eigenvalues.
From the plot we can see that our proposed method SDAF has the highest number of significant eigenvalues (much greater than zero) among the three methods, in which containing more significant eigenvalues means that the corresponding representations have large variance along numerous orthogonal directions. Thus, the proposed SDAF captures diverse features on the same dataset, contributing to benefit the learning on the future tasks. Note that from 40th eigenvalues they are nearly zero, so we skip those for a clear plot.
}


\walon{
\subsubsection{Execution Time}
Figure \ref{fig:time} shows the execution time (including training and inference) of the baselines and our method, based on the experiments in CIFAR-10 with a memory size $\mathcal{M} = 500$. We ran the experiment on a computer equipped with a Xeon E5-2620 CPU and a GTX-1080ti GPU.
The proposed method spends less executive time since the number of total training iterations is shorter than others though the SDA multiplies the mini-batch $K$ by times which can be easily accelerated by GPU parallelism.
}


\begin{figure*}[htpb]
     \centering
     \includegraphics[scale=0.6]{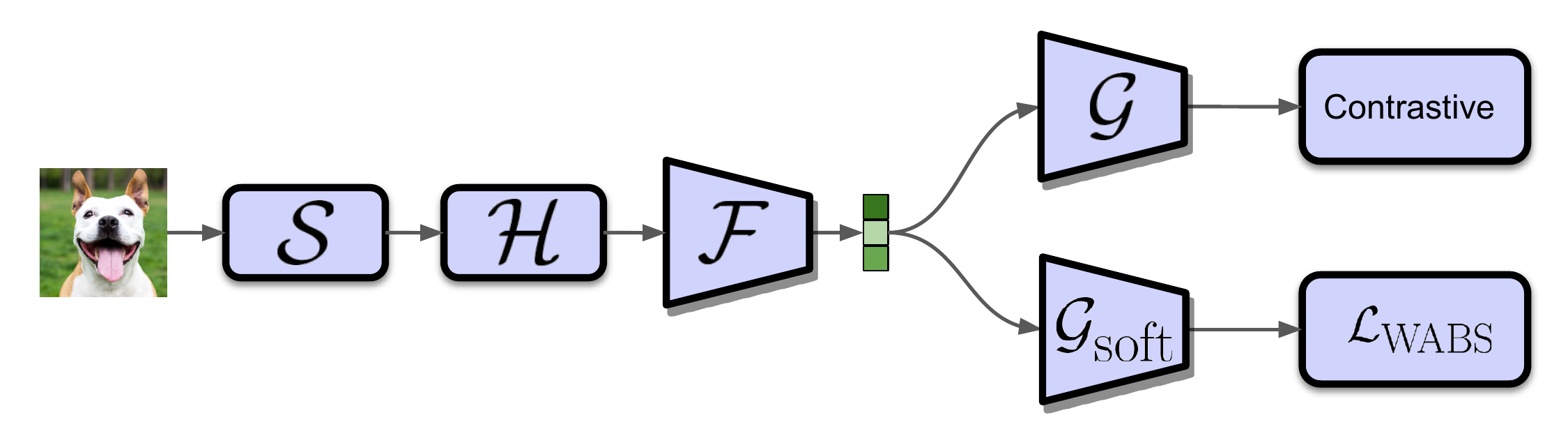}
     \caption{\walon{Model architecture. $\mathcal{S}$ is semantically distinct augmentation. $\mathcal{H}$ is the transformation module for contrastive learning which is described in Section \ref{detail_parameters}. $\mathcal{F}$ is a feature extractor. $\mathcal{G}$ is a multilayer perceptron (MLP). 
     $\mathcal{G}_{\text{soft}}$ is a single-layer softmax classifier (i.e. built upon one fully-connected layer).}}
     \label{fig:architecture}
\end{figure*}

\walon{
\subsection{Network Architecture}
The full architecture of our proposed SDAF model is illustrated in Figure~\ref{fig:architecture}. First, the semantically distinct augmentation $\mathcal{S}$ generates $K$ different images $S_k(x)$ from the input image $x$. Then, we sample $2K$ random transformations $H \sim \mathcal{H}$ to create $2K$ views for those $K$ images, followed by using the feature extractor $\mathcal{F}$ to project the $2K$ images into the latent representation space.
The network $\mathcal{G}$ further projects the resultant feature representation into another low-dimensional space to perform the contrastive learning. On the other hand, the softmax classifier $\mathcal{G}_{\text{soft}}$ is responsible for computing the cross-entropy loss, which is equipped with the weight-aware balanced sampling strategy ($\mathcal{L}_{\text{WABS}}$) upon the representations.
}

\subsection{Implementation Details} \label{detail_parameters}
\subsubsection{Hyperparameters}
The network architecture for the components used in our proposed method basically follows the ones in~\cite{mai2021scr}.
For all experiments, we adopt a reduced ResNet18 \cite{he2016resnet} as our feature extractor $\mathcal{F}$ with resultant feature dimension set to 160, the projection head $\mathcal{G}$ is a two-layer multilayer perceptron (MLP) with width 160 and 128 respectively, \iclr{and the predictor $\mathcal{P}$ is also a two-layer MLP with both input and output width being 128}. The transformation module $\mathcal{H}$ of contrastive learning consists of random cropping, \iclr{random horizontal flip}, random colour distortion, and random grey scale. The detailed setting for transformation module $\mathcal{H}$ is described later in the next paragraph. We adopt the SGD optimiser with learning rate \iclr{0.1}. The batch size $\left|B^u_t\right|$ is 10, and the retrieval batch size $\left|B_M\right|$ is 10.
We adopt the \iclr{rotation as the SDA strategy $\mathcal{S}$, where $\mathcal{S}$ consists of four different degree of rotation ($K=4$), i.e. $0^{\circ}, 90^{\circ}, 180^{\circ},$ and $270^{\circ}$}. We empirically set $\lambda$ in the total loss function (cf. Eq.~\ref{eq:total_loss}) to $1.5$, and the temperature $\tau_w$ in our WABS (cf. Eq.~\ref{eq:rate}) to 0.5. 
We use reservoir sampling \cite{10.1145/3147.3165reservoir} and uniform random sampling for operations MemoryUpdate and MemoryRetrieval as described in Algorithm~\ref{alg:algorithm}.

\subsubsection{Transformation Module} \label{transformation_module}
\iclr{
The transformation module $\mathcal{H}$ for contrastive learning composes of a long series of random transformations and another short series of random transformations, where each series is responsible for generating one view in a positive pair (following the common practice as used in FixMatch~\cite{fixmatch}).
The long series of random transformations includes a uniformly random cropping, a random horizontal flip, a random colour distortion, and a random grey scale operator. 
Uniformly random cropping keeps the original image from 50\% to 100\% in terms of area, and it has been implemented in Pytorch~\cite{pytorch} as ``torchvision.transforms.RandomResizedCrop''.
Random horizontal flip has 50\% chance to flip the image.
Random colour distortion has also been realized in Pytorch~\cite{pytorch} as ``torchvision.transforms.ColorJitter'', and we set the factors for brightness, contrast, saturation to $0.4$, and the factor for hue is set to $0.1$.
Last, with a chance of 20\%, the image is converted to gray scale, and this operator is implemented as ``torchvision.transforms.RandomGrayscale'' in Pytorch~\cite{pytorch}.
On the other hand, the short series of random transformations comprises of a uniformly random cropping and a random colour distortion. The implementation is the same as the long series of random transformation, except the cropping area is set to 75\% to 100\%, the factors for brightness, contrast, saturation to $0.2$, and the factor for hue is set to $0.05$.
}


\end{document}